\documentclass{article}%
\usepackage{amsmath}
\usepackage{amsfonts}
\usepackage{amssymb}
\usepackage{graphicx}%
\providecommand{\U}[1]{\protect\rule{.1in}{.1in}}
\begin{document}

\title{Function Trees: Transparent Machine Learning}
\author{Jerome H. Friedman$\thanks{Department of Statistics, Stanford University,
Stanford, CA 94305, USA. (jhf@stanford.edu)}$}
\maketitle

\begin{abstract}
The output of a machine learning algorithm can usually be represented by one
or more multivariate functions of its input variables. Knowing the global
properties of such functions can help in understanding the system that
produced the data as well as interpreting and explaining corresponding model
predictions. A method is presented for representing a general multivariate
function as a tree of simpler functions. This tree exposes the global internal
structure of the function by uncovering and describing the combined joint
influences of subsets of its input variables. Given the inputs and
corresponding function values, a function tree is constructed that can be used
to rapidly identify and compute all of the function's main and interaction
effects up to high order. Interaction effects involving up to four variables
are graphically visualized.

\end{abstract}

\section{Introduction}

A fundamental exercise in machine learning is the approximation of a function
of several to many variables given values of the function, often contaminated
with noise, at observed joint values of the input variables. The result can
then be used to estimate unknown function values given corresponding inputs.
The goal is to accurately estimate the underlying (non noisy) outcome values
since the noise is by definition unpredictable. To the extent that this is
successful the estimated function may, in addition, be used to try to
understand underlying phenomena giving rise to the data. Even when prediction
accuracy is the dominate concern, being able to comprehend the way in which
the input variables are jointly combining to produce predictions may lead to
important sanity checks on the validity of the function estimate. Besides
accuracy, the success of this latter exercise requires that the structure of
the function estimate be represented in a comprehensible form.

It is well known, and often commented, that the most accurate function
approximation methods to date tend not to provide comprehensible results. The
function estimate is encoded in an obtuse form that obscures potentially
recognizable relationships among the inputs that give rise to various function
output values. This is especially the case when the function is not inherently
low dimensional. That is, medium to large subsets of the input variables act
together to influence the function $F(\mathbf{x})$ in the sense that their
combined contribution cannot be represented by a combination of smaller
subsets of those variables (interaction effects). This has led to incomplete
interpretations based on low dimensional approximations such as additive
modeling with no interactions, or models restricted to at most two--variable interactions.

\section{Interaction effects}

For input variables $\mathbf{x}\,=$ $(x_{1},x_{2},\cdot\cdot\cdot,x_{p})$ a
function $F(\mathbf{x})$ is said to exhibit an interaction between two of them
$x_{j}$ and $x_{k}$ if the difference in the value of $F(\mathbf{x})$ as
result of changing the value of one of them $x_{j}$ depends on the value of
the other $x_{k}$. This means that in order to understand the effect of these
two variables on $F(\mathbf{x})$ they must be considered together and cannot
be studied separately. An interaction effect between variables $x_{j}$ and
$x_{k}$ implies that the second derivative of $F(\mathbf{x})$ jointly with
respect to $x_{j}$ and $x_{k}$ is not zero for at least some values of
$\mathbf{x}$. That is,%
\begin{equation}
E_{\mathbf{x}}\left[  \frac{\partial^{2}F(\mathbf{x})}{\partial x_{j}\partial
x_{k}}\right]  ^{2}>0\text{,} \label{e1}%
\end{equation}
with an analogous expression involving finite differences for categorical
variables. If there is no interaction between these variables, the function
$F(\mathbf{x})$ can be expressed as a sum of two functions, one that does not
depend on $x_{j}$ and the other that is independent of $x_{k}$%
\begin{equation}
F(\mathbf{x})=f_{\backslash j}(\mathbf{x}_{\backslash j})+f_{\backslash
k}(\mathbf{x}_{\backslash k})\text{.} \label{e2}%
\end{equation}
Here $\mathbf{x}_{\backslash j}$ and $\mathbf{x}_{\backslash k}$ respectively
represent all variables except $x_{j}$ and $x_{k}$. If a given variable
$x_{j}$ interacts with no other variable, then the function can be expressed
as%
\begin{equation}
F(\mathbf{x})=f_{j}(x_{j})+f_{\backslash j}(\mathbf{x}_{\backslash j})
\label{e2.1}%
\end{equation}
where the first term on the right is a function only of $x_{j}$ and the second
is independent of $x_{j}$. In this case $F(\mathbf{x})$ is said to be
\textquotedblleft additive\textquotedblright\ in variable $x_{j}$ and the
univariate function $f_{j}(x_{j})$ can be examined to study the effect of
$x_{j}$ on the function $F(\mathbf{x})$ independently from the effects of the
other variables.

Higher order interactions are analogously defined. A function $F(\mathbf{x})$
is said to have an $n$--variable interaction \ effect involving variables
$\{x_{j}\}_{1}^{n}$ if%
\begin{equation}
E_{\mathbf{x}}\left[  \frac{\partial^{n}F(\mathbf{x})}{\partial x_{1}\partial
x_{2}\cdot\cdot\cdot\;\partial x_{n}}\right]  ^{2}>0\text{,}\label{e3}%
\end{equation}
again with an analogous expression involving finite differences for
categorical variables. The existence of such an $n$--variable interaction
implies that the effect of the corresponding variables $\{x_{j}\}_{1}^{n}$ on
the function $F(\mathbf{x})$ cannot be decomposed into a sum of functions each
involving subsets of those variables. If there is no such interaction, the
contribution of variables $\{x_{j}\}_{1}^{n}$ to the variation of
$F(\mathbf{x})$ can be decomposed into a sum of functions each not depending
upon one of these respective variables%
\begin{equation}
F(\mathbf{x})=\sum_{j=1}^{n}f_{\backslash j}(\mathbf{x}_{\backslash
j})\text{.}\label{e3.1}%
\end{equation}
If none of the variables in a subset $s\mathbf{=}\{x_{j}\}_{1}^{n}$ interact
with any of the variables in its complement set $\backslash s\mathbf{=}%
\{x_{j}\}_{n+1}^{p}$, then the function is additive in the variable subset $s$%
\begin{equation}
F(\mathbf{x})=f_{s}(\mathbf{x}_{s})+f_{\mathbf{\backslash s}}(\mathbf{x}%
_{\backslash s})\label{e3.2}%
\end{equation}
and one can study the effect of $\{x_{j}\}_{1}^{n}$ on the function
$F(\mathbf{x})$ separately from that of the other variables.

The notion of interaction effects can be useful for understanding a function
$F(\mathbf{x})$ if it is dominated by those of low order involving variable
subsets $s\subset\{x_{j}\}_{1}^{p}$ with small cardinality ($|\,s\,|\,<<p$).
In this case a target function of many variables can be understood in terms of
a collection of lower dimensional functions each depending on a relatively
small different subset of the variables. It is generally easier to comprehend
the lower dimensional structure associated with fewer variables. Also, a
common regularization method is to more heavily penalize or forbid solutions
involving higher order interactions. Although often useful, this can result in
lower accuracy and/or misleading interpretations when $F(\mathbf{x})$ involves
substantial interaction effects of higher order than that assumed. As seen
below the existence of such higher order interactions can mislead
interpretation by causing the functional form of those of lower order to
depend on values of other unknown variables.

\begin{center}
$%
\begin{array}
[c]{cc}%
{\parbox[b]{3.0718in}{\begin{center}
\includegraphics[
height=3.1479in,
width=3.0718in
]%
{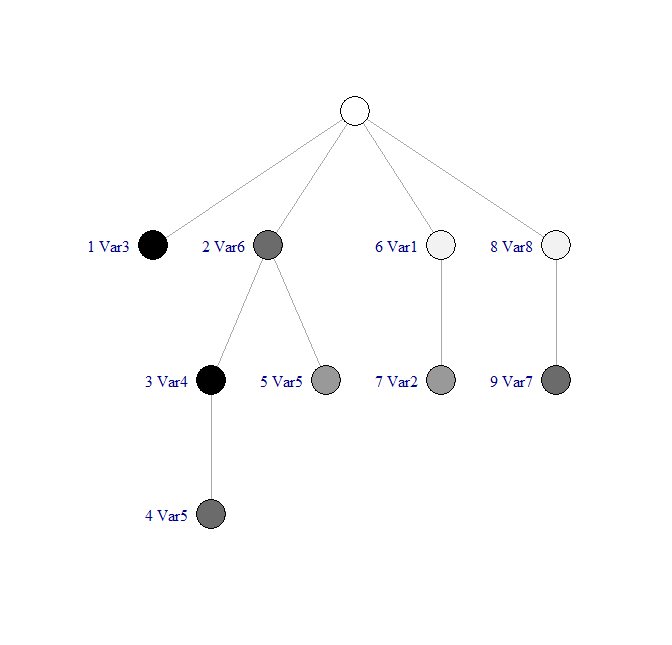}%
\\
Figure 1: Function tree.
\end{center}}}
&
{\parbox[b]{2.3877in}{\begin{center}
\includegraphics[
height=2.9568in,
width=2.3877in
]%
{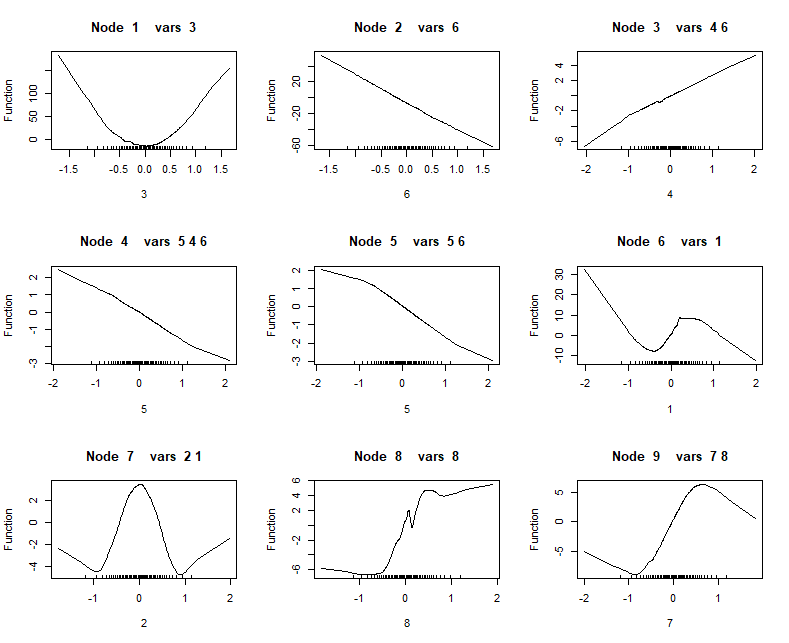}%
\\
Figure 2: Node functions.
\end{center}}}
\end{array}
$
\end{center}

\section{Function trees}

A function tree is a way of representing a multivariate function so as to
uncover and estimate its interaction effects of various orders. The input
function $F(\mathbf{x})$ is defined by data $\{\mathbf{x}_{i},y_{i}\}_{1}^{N}$
with $\mathbf{x}$ representing multivariate evaluation points and $y$ the
corresponding function values perhaps contaminated with noise. The output of
the procedure is a set of univariate functions, each one depending on a single
selected input variable. These functions are arranged as the non root nodes of
a tree. A constant function is associated with the root. The tree provides a
prescription for combining these univariate functions to form the multivariate
function approximation $\hat{F}(\mathbf{x})$.

Figure 1 shows one such function tree derived from simple simulated data for
illustration. Figure 2 shows the nine non constant functions corresponding to
each of the respective non root nodes of the tree. In addition to its
associated univariate function (Fig. 2), each tree node $k$ represents a basis
function $B_{k}(\mathbf{x})$. The sum of these basis functions forms the final
multivariate approximation%
\begin{equation}
\hat{F}_{K}(\mathbf{x})=B_{0}+\sum_{k=1}^{K}B_{k}(\mathbf{x})\text{.}
\label{e4}%
\end{equation}
Each node's basis function is the product of its associated univariate
function and those of the nodes on the path from it to the root%
\begin{equation}
B_{k}(\mathbf{x})=%
%TCIMACRO{\dprod \limits_{l\in p(k)}}%
%BeginExpansion
{\displaystyle\prod\limits_{l\in p(k)}}
%EndExpansion
f_{l}(x_{j(l)})\text{.} \label{e5}%
\end{equation}
Here $p(k)$ represents those nodes $l$ on the path from node $k$ to the root
and $j(l)$ labels the input variable associated with each such node $l$. The
darkness of each node's shading in Fig. 1 corresponds to the estimated
relative influence of its basis function on the final function estimate as
measured by its standard deviation over the training data.

\setcounter{figure}{2} The interaction level of each basis function (\ref{e5})
is the number of univariate functions of \emph{different} variables along the
path from its corresponding node to the root of the tree. Products of
univariate functions involving $n$ different variables satisfy (\ref{e3}).
Note that different univariate functions of the \emph{same} variable may
appear multiple times along the same path. While potentially improving the
model fit they do not increase interaction level. With this approach
interactions among a specified subset of the variables $s$ are modeled by sums
of products of univariate functions of those variables%
\begin{equation}
J_{s}(\{x_{j}\}_{j\in s})=\sum_{k=1}^{K_{s}}%
%TCIMACRO{\dprod \limits_{j\in s}}%
%BeginExpansion
{\displaystyle\prod\limits_{j\in s}}
%EndExpansion
f_{jk}(x_{j})\text{.} \label{e6}%
\end{equation}

The function tree model shown in Figs. 1 \& 2 was obtained by applying the
procedure described below to a simulated data example. There are $10000$
observations with outcome variables generated as $y=F(\mathbf{x})+\varepsilon$
with $\mathbf{x}\sim N^{8}(0,0.5)$ and%
\begin{equation}
F(\mathbf{x})=4\,\sin(\pi x_{1})\,\cos(\pi x_{2})+7\,\,x_{3}^{2}%
+15\,\,(x_{4}+0.4)\,(x_{5}-0.6)\,(x_{6}+0.2)+5\,\sin(\pi\,(x_{7}%
+0.1)\,x_{8})\text{.} \label{e77}%
\end{equation}
\linebreak The noise is generated as $\varepsilon\sim N(0,var(F)/4)$ producing
a $2/1$ signal/noise ratio. This target function has an additive dependence on
the third variable $x_{3}$, separate two--variable interactions between
$(x_{1},x_{2})$ and $(x_{7},x_{8})$ respectively, and a trilinear
three--variable interaction among $(x_{4},x_{5},x_{6})$.

The first node of the tree is seen to incorporate the pure additive effect of
$x_{3}$. Nodes 2 - 5 model the $(x_{4},x_{5},x_{6})$ three--variable
interaction. The two--variable interaction between $x_{1}$ and $x_{2}$ is
modeled by nodes 6 and 7, and that of $x_{7}$ and $x_{8}$ is modeled by nodes
8 and 9. The nine univariate functions (Fig. 2) combined as specified by the
tree (Fig. 1) produce a multivariate function that explains 97\% of the
variance of the target (\ref{e77}).

\subsection{Tree construction\label{con}}

Function trees are constructed in a standard forward stepwise best first
manner. Initially the tree consists of a single root node representing a
constant function $B_{0}$. At the $K$th step there are $K$ basis functions
(\ref{e4}) (\ref{e5}) in the current model $\hat{F}_{K}(\mathbf{x})$. The
($K+1)$st is taken to be
\begin{equation}
B_{K+1}(\mathbf{x})=B_{k^{\ast}}(\mathbf{x})\,f^{\ast}(x_{j^{\ast}})
\label{e8}%
\end{equation}
where $j^{\ast}$, $k^{\ast}$, and $f^{\ast}\,$are the solution to%
\begin{equation}
(j^{\ast},k^{\ast},\,f^{\ast})=\arg\min_{(j,k,\,\,f)}\,\hat{E}\left(
y-\hat{F}_{K}(\mathbf{x})-B_{k}(\mathbf{x})\,f(x_{j})\right)  ^{2}\text{.}
\label{e9}%
\end{equation}
Here $B_{k}(\mathbf{x})$ is one of the basis functions in the current model,
$f(x_{j})$ is a function of one of the predictor variables $x_{j}$ and the
empirical expected value is over the data distribution. The model is then
updated%
\begin{equation}
\hat{F}_{K+1}(\mathbf{x})=\hat{F}_{K}(\mathbf{x})+B_{k^{\ast}}(\mathbf{x}%
)\,f^{\ast}(x_{j^{\ast}}) \label{e10}%
\end{equation}
for the next iteration. In terms of tree construction the update consists of
adding a daughter node to current node $k^{\ast}$ with associated function
$\,f^{\ast}(x_{j^{\ast}})$. Iterations continue until the goodness--of--fit
stops improving as measured on an independent test data sample.

\subsection{Backfitting\label{bak}}

As with all tree based methods, smaller trees (less nodes) tend to be more
easily understood. Here tree size can be reduced and accuracy improved by
increased function optimization (backfitting) at each step (Friedman and
Stuetzle 1981). This is implemented here by updating the functions associated
with all current tree nodes in presence of the newly added one, and those
previously updated. In particular, after adding the new ($K+1$)st node, all
node functions are re-estimated $f_{k}(x_{j(k)})\leftarrow f_{k}^{\ast
}(x_{j(k)})$ one at a time in order starting from the first
\begin{equation}
f_{k}^{\ast}(x_{j(k)})=\arg\min_{\tilde{f}}\,\hat{E}_{\mathbf{x}}\left[
y-\hat{F}_{K+1}(\mathbf{x})-B_{k}(\mathbf{x})\left(  \tilde{f}(x_{j(k)}%
)/\,f_{k}(x_{j(k)})-1\right)  \right]  ^{2}\text{,} \label{e11}%
\end{equation}%
\begin{equation}
\hat{F}_{K+1}(\mathbf{x})\leftarrow\hat{F}_{K+1}(\mathbf{x})+B_{k}%
(\mathbf{x})\left(  f^{\ast}(x_{j(k)})/f_{k}(x_{j(k)})-1\right)  \text{,
}k=1,2,\cdot\cdot\cdot,(K+1)\text{.} \label{e12}%
\end{equation}
Here $k$ labels a node of the tree and $f_{k}$ its associated function of
variable $x_{j(k)}$. In terms of tree construction, each such step involves
replacing each ($k$th) node's current function $f_{k}(x_{j(k)})$ with
$f_{k}^{\ast}(x_{j(k)})$ (\ref{e11}). No changes are made to tree topology
(Fig. 1) or the input variables associated with each node. Only the functions
(Fig. 2) are updated. Backfitting passes can be repeated until the model
becomes stable. This usually happens after one or two passes. Note that in the
presence of backfitting function tree models are not nested. All of the basis
functions of the $K+1$ node model are different from those of its $K$ node predecessor.

\subsection{Univariate function estimation}

The central operation of the above procedure is repeated estimation of
univariate functions of the individual predictor variables. In both (\ref{e9})
and (\ref{e11}) the solutions take the form of weighted conditional
expectations%
\begin{equation}
\hat{f}(x_{j})=\hat{E}_{w^{2}}\left[  \frac{r}{w}\,|\,x_{j}\right]  \text{,}
\label{e13}%
\end{equation}
where $w$ represents one of the current basis functions $w=B_{k}(\mathbf{x})$
and $r=y-\hat{F}(\mathbf{x})$ are the current residuals. The function tree
procedure as described above is agnostic concerning the methods used for this
estimation. However, actual performance in terms of execution speed and
prediction accuracy can be highly dependent on such choices.

A major consideration in the choice of an estimation method for a particular
variable $x_{j}$ is the nature of that variable in terms of the values it
realizes, and connections between those values. A categorical variable
(factor) realizes discrete values with no order relation. The only method for
evaluating (\ref{e13}) for such variables is to take the weighted ($w^{2}$)
mean of $r/w$ at each discrete $x$ value. This is also a viable strategy if
the $x_{j}$ values are orderable but realize only a small set of distinct values.

For numeric variables that realize many distinct orderable values one can take
advantage of the presumed smoothness of the solution function to improve
accuracy by borrowing strength from similarly valued observations. There are
many such methods for \textquotedblleft smoothing\textquotedblright\ data (see
Irizarry 2019). For the examples presented below near neighbor local averaging
and local linear fitting were employed. Near neighbor local averaging
estimates are equivariant under monotone transformations of the $x$-values,
depending only their relative ranks. This can provide higher accuracy in the
presence of irregular or clumped $x$-values, immunity to $x$ outliers,
resistance to $y$ outliers and more cautious (constant) extrapolation at the
edges. For more evenly distributed data with many distinct $x$-values, in the
absence of $x$ and $y$ outliers, local linear fitting can yield smoother more
accurate results.

\section{Interpretation}

The primary goal of the function tree representation is to enhance
interpretation by exposing a function's separate interaction effects of
various orders. These can then be studied using graphical and other methods to
gain insight into the nature of the relationship between the predictor
variables $\mathbf{x}=(x_{1},x_{2},\cdot\cdot\cdot,x_{p})$ and \ the target
function $F(\mathbf{x})$.

\subsection{Partial dependence functions\label{PDF}}

One way to estimate the contribution of subsets of predictor variables to a
function $F(\mathbf{x})$\ is through partial dependence functions (Friedman
2001). Let $\mathbf{z}$ represent a subset of the predictor variables
$\mathbf{z}\subset\mathbf{x}$ and $\mathbf{\tilde{z}}$ the complement subset
$\mathbf{z\,}\cup\,\mathbf{\tilde{z}\,}\mathbf{=x\,}$. Then the partial
dependence of a function \thinspace$F(\mathbf{x})\,\ $on $\mathbf{z}$ is
defined as%
\begin{equation}
PD(\mathbf{z})=E_{\mathbf{\tilde{z}}}\left[  F(\mathbf{x})\right]  \text{.}
\label{e14}%
\end{equation}
Conditioned on joint values of the variables in $\mathbf{z}$, the the value of
the function $F(\mathbf{x})$ is averaged over the joint values of the
complement variables $\mathbf{\tilde{z}}$. Since the joint locations of
partial dependence functions are generally not identifiable, they are each
centered to have zero mean value over the data distribution of $\mathbf{x}$.

If for a function $F(\mathbf{x})$ the variables in the specified subset
$\mathbf{z}$ do not participate in interactions with variables in
$\mathbf{\tilde{z}}$ then the additive dependence (\ref{e3.2}) of
\ $F(\mathbf{x})$ on $\mathbf{z}$ is well defined and given by $PD(\mathbf{z}%
)$. If this is not the case, then the dependence of the target function
$F(\mathbf{x})$ on the subset $\mathbf{z}$ is not well defined in the sense
that its functional form changes with changing values of the variables in
$\mathbf{\tilde{z}}$. In this case one can define a \textquotedblleft
nominal\textquotedblright\ dependence by averaging over some distribution
$p(\mathbf{x})$ of the predictor variables $\mathbf{x}$. This confounds the
properties of the actual target function $F(\mathbf{x})$ with those of the
data distribution $p(\mathbf{x})$ on the resulting estimate of the dependence
on $\mathbf{z}$. Two popular choices for $p(\mathbf{x})$ are the training data
and the product of its marginals (independence).\ Partial dependence functions
choose a compromise in which the variables in $\mathbf{z}$ are taken to be
independent of those in $\mathbf{\tilde{z}}$. Note that the result between
this and using the training data distribution is different only if there are
substantial correlations or associations between the interacting variables in
$\mathbf{z}$ and $\mathbf{\tilde{z}}$.

Partial dependence functions (\ref{e14}) can be estimated from the data in a
straightforward way by evaluating%
\begin{equation}
\widehat{PD}(\mathbf{z})=\frac{1}{N}\sum_{i=1}^{N}F(\mathbf{z,\tilde{z}}_{i})
\label{e15}%
\end{equation}
over a representative set of joint values of $\mathbf{z}$. This requires
$N\cdot N_{\mathbf{z}}$ target function evaluations where $N_{\mathbf{z}}$ is
the number of evaluation points and $N$ is the sample size used for averaging.

With the function tree representation partial dependence functions can be
computed much more rapidly. For the purpose of computing a partial dependence
function on a variable subset $\mathbf{z}$ the function tree model can be
expressed as%
\begin{equation}
\hat{F}(\mathbf{x})=A+\sum_{k=1}^{K}f_{k}(\mathbf{z)\cdot\,}g_{k}%
\mathbf{(\tilde{z})} \label{e16}%
\end{equation}
where $A$ represents all basis functions not involving any variables in
$\mathbf{z}$, $\,f_{k}(\mathbf{z)}$ are functions involving variables only in
$\mathbf{z}$, and $\mathbf{\tilde{z}}$ represents the compliment variables.
With this representation the partial dependence of $\hat{F}(\mathbf{x})$ on
the variables $\mathbf{z}$ is simply%
\begin{equation}
\widehat{PD}(\mathbf{z})=\sum_{k=1}^{K}\bar{g}_{k}\cdot f_{k}(\mathbf{z)}
\label{e17}%
\end{equation}
where $\bar{g}_{k}=\hat{E}_{\mathbf{x}}[g_{k}\mathbf{(\tilde{z})]}$ is the
mean of $g_{k}\mathbf{(\tilde{z})}$ over the data. This (\ref{e17}) is a
particular linear combination of the functions $\{f_{k}(\mathbf{z)\}}_{1}^{K}%
$. The number of target function $\hat{F}(\mathbf{x})$ evaluations required to
compute (\ref{e17}) is proportional to
\begin{equation}
C(N,N_{\mathbf{z}})=N_{\mathbf{z}}\,+\alpha(\mathbf{z,\tilde{z}})\cdot N
\label{e17.1}%
\end{equation}
where $\alpha(\mathbf{z,\tilde{z}})$ is the fraction of node basis functions
(\ref{e5}) involving variables in both $\mathbf{z}$ and $\mathbf{\tilde{z}}$.

\subsection{Partial association functions\label{PAF}}

Partial dependence functions focus on the properties of the function
$F(\mathbf{x})$ by averaging over a distribution in which the $\mathbf{z}$
variables are independent of those in $\mathbf{\tilde{z}}$. This concentrates
on the target function by removing the effects of associations between those
variable subsets. As a result the calculation can sometimes emphasize
$\mathbf{x}$-values for which the data density $p(\mathbf{x})$ is small
leading to potential inaccuracy. This only affects a partial dependence to the
extent that there are variables in $\mathbf{\tilde{z}}$ with which its
variables $\mathbf{z}$ both interact and are correlated.

The function tree representation provides a method for detecting and measuring
the strength of such occurrences. \textquotedblleft Partial
association\textquotedblright\ functions are defined as%
\begin{equation}
\widehat{PA}(\mathbf{z})=\sum_{k=1}^{K}f_{k}(\mathbf{z)}\cdot h_{k}%
(\mathbf{z}) \label{e18}%
\end{equation}
where%
\begin{equation}
h_{k}(\mathbf{z})=E_{\mathbf{x}}\left[  g_{k}\mathbf{(\tilde{z})\,|\,\,}%
f_{k}(\mathbf{z)}\right]  \label{e19}%
\end{equation}
with the functions $f_{k}(\mathbf{z)}$ and $g_{k}\mathbf{(\tilde{z})}$ defined
in (\ref{e16}). As with partial dependence functions (\ref{e17}) this can be
viewed as a linear combination of the functions $\{f_{k}(\mathbf{z)\}}_{1}%
^{K}$ but with varying coefficients $\{h_{k}(\mathbf{z)\}}_{1}^{K}$ that
account for the associations between the variables in $\mathbf{z}$ and the
complement variables $\mathbf{\tilde{z}}$. The coefficient functions
$h_{k}(\mathbf{z})$ can be estimated by any univariate smoother. Regression
splines with knots at the $20$ percentiles are used in the examples below.
Note that from (\ref{e16}) partial association functions (\ref{e18})
(\ref{e19}) reduce to partial dependence functions (\ref{e17}) when variables
in $\mathbf{z}$ do not participate in interactions with variables in
$\mathbf{\tilde{z}\,\ }$($\,g_{k}\mathbf{(\tilde{z}})=1$), or they are
independent of those in $\mathbf{\tilde{z}\,\ }$with which they do interact
($\,h_{k}\mathbf{(z})=\bar{g}$). Otherwise they can produce different results
caused by the variation in the respective coefficient functions $h_{k}%
(\mathbf{z)}$ induced by the associations between the corresponding
interacting variables in $\mathbf{z}$ and $\mathbf{\tilde{z}}$.

Partial association functions can be used to access the influence that
associations among the predictor variables have on the corresponding partial
dependences. In particular, one can substitute them (\ref{e18}) (\ref{e19})
for partial dependence functions (\ref{e17}) in any analysis. Similar results
indicate that such influences are small.

\subsection{ Interaction detection}

Partial dependence functions can be used to detect interaction effects between
variables. For example, if a target function $F(\mathbf{x})$ contains no
interaction between variables $x_{j}$ and $x_{k}$ its partial dependence on
those variables is%
\begin{equation}
PD(x_{j},x_{k})=PD(x_{j})+PD(x_{k}) \label{e20}%
\end{equation}
(Friedman and Popescu 2008). If there is such an interaction, $PD(x_{j}%
)+PD(x_{k})$ represents the additive component and
\begin{equation}
I(x_{j},x_{k})=PD(x_{j},x_{k})-PD(x_{j})-PD(x_{k}) \label{e21}%
\end{equation}
represents the corresponding pure interaction component of the effect of
$x_{j}$ and $x_{k}$ on $F(\mathbf{x})$ as reflected by its partial dependences.

More generally, the pure interaction effect involving variables $s=\{x_{j}%
\}_{1}^{n}$ is defined to be
\begin{equation}
I(s)=PD(s)-\sum_{u\subset s}I(u) \label{e22}%
\end{equation}
where $I(u)$ is recursively defined as%
\begin{equation}
I(u)=PD(u)-\sum_{v\subset u}I(v)\text{.} \label{e23}%
\end{equation}
This pure interaction component $I(\{x_{j}\}_{1}^{n})$\ is its corresponding
partial dependence with all lower order interactions among all its variable
subsets removed. If there is no $n$--variable interaction effect involving
$\{x_{j}\}_{1}^{n}$ its pure interaction component $I(\{x_{j}\}_{1}^{n})=0$.
Note that this result depends only on the properties of the function
$F(\mathbf{x})$ (\ref{e3.1}) and not on the data distribution $p(\mathbf{x})$.
The strength of such an interaction effect can be taken to be the standard
deviation of its pure interaction component over the training data, or other
specified data distribution, divided by the standard deviation of the
corresponding target function
\begin{equation}
S(\{j\}_{1}^{n})=\sqrt{\left.  var_{\mathbf{x}}[I(\{x_{j}\}_{1}^{n})]\right/
var_{\mathbf{x}}(F(\mathbf{x}))}\text{.} \label{e24}%
\end{equation}

If there exists a higher order interaction effect jointly involving variables
in subset $\{x_{j}\}_{1}^{m}$ along with other variables $\{x_{j}\}_{m+1}^{n}%
$, $S(\{j\}_{1}^{n})>0$ (\ref{e24}), then the form of the interaction function
of the subset $I(\{x_{j}\}_{1}^{m}\,|\,\{x_{j}\}_{m+1}^{n})$ depends on the
joint values of the complement variables $\{x_{j}\}_{m+1}^{n}$. The resulting
unconditioned interaction effect $I(\{x_{j}\}_{1}^{m})$ is then an average
over the joint distribution $p(\{x_{j}\}_{m+1}^{n})$ of the complement
variables. If there are no such higher order interactions involving
$\{x_{j}\}_{1}^{m}$ jointly with other variables, its pure interaction effect
$I(\{x_{j}\}_{1}^{m})$ does not depend on the data distribution $p(\mathbf{x}%
)$ and reflects only properties of the target function $F(\mathbf{x})$.

The measure $S(s)$ (\ref{e24}) indicates the strength of an $n$--variable
interaction effect involving the corresponding variable subset $s=$
$\{x_{j}\}_{1}^{n}$ over the distribution $p(\mathbf{x})$. One can search for
substantial interaction effects by evaluating (\ref{e22}--\ref{e24}) over all
variable subsets $s$ up to some maximum size $|\,s\,|\,\leq M$. This approach
can in principle be applied to a target function $F(\mathbf{x})$ represented
in any form. All that is required to compute partial dependence functions is
the value of the function at various specified values of $\mathbf{x}$.%

\begin{figure}[ptb]%
\centering
\includegraphics[
height=3.7421in,
width=4.5247in
]%
{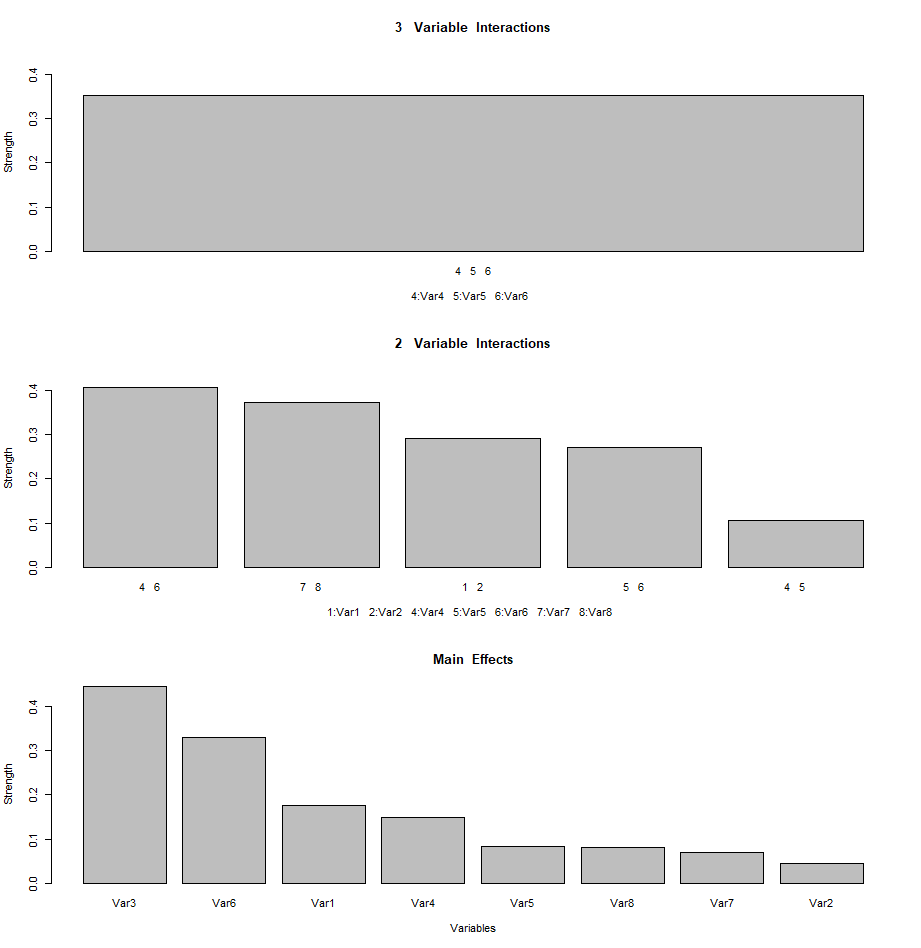}%
\caption{Strengths of indentifed main and interaction effects in the synthetic
data example \ of Section 3.}%
\label{fig4}%
\end{figure}

Figure \ref{fig4} shows the strengths (\ref{e24}) of all interaction effects
uncovered in the synthetic data example of Section 3 using this approach. The
presence of more lower order interaction effects shown here than explicitly
represented by the function tree (Fig. 1) is due to lower level marginals of
higher order basis functions (\ref{e6}) being assigned to their proper
interaction level.

\section{Computation}

For a total of $p$ variables the number of distinct subsets involving $n<p$
variables is $\binom{p}{n}$. For each such subset the number of partial
dependence functions required to extract its pure interaction effect
(\ref{e22}) (\ref{e23}) grows exponentially in its size $n.$ Thus for larger
values of $p$ and $n$ the required computation using the standard method
(\ref{e15}) to evaluate all of the necessary partial dependence functions is
generally too severe to allow a search for interaction effects by complete
enumeration over all variable subsets. However, for functions represented by a
function tree, one can employ (\ref{e17}) to dramatically reduce the
computation (\ref{e17.1}) required to evaluate each partial dependence
function. This in turn allows the search for interactions to be performed over
many more and larger variable subsets representing higher order interaction effects.

For small to moderate number of variables $p\lesssim20$ complete enumeration
using (\ref{e17}) is generally feasible for $n\lesssim4$. However, for larger
problems the rapid growth with respect to both $p$ and $n$ soon places the
required computation out of reach. Parallel computation and smart strategies
for reusing previously computed partial dependence functions can somewhat ease
this burden. However, the properties of partial dependence functions allow a
simple input variable screening approach that can often considerably reduce
the size of the search.

If for a (centered) function $F(\mathbf{x})$ a variable $x_{j}$ participates
in no interactions with any other variables then it can be expressed as%
\begin{equation}
F(\mathbf{x})=PD(x_{j})+PD(\mathbf{x}_{\backslash j}) \label{e33}%
\end{equation}
where $PD(x_{j})$ is its partial dependence on $x_{j}$ and $PD(\mathbf{x}%
_{\backslash j})$ is its partial dependence on all other variables (Friedman
and Popescu 2008). One can define an overall interaction strength for each
predictor variable as%
\begin{equation}
H_{j}=\sqrt{\hat{E}_{\mathbf{x}}(F(\mathbf{x})-PD(x_{j})-PD(\mathbf{x}%
_{\backslash j}))^{2}}\text{.} \label{e34}%
\end{equation}
Variables $x_{j}$ with small values of $H_{j}$ can be removed from the search
for interaction effects. This often excludes many variables thereby
substantially reducing computation.

Computation can be further reduced by taking advantage of the function tree
representation. The function tree strength of the contribution of input
variable $x_{j}$ to interaction level $k$ is taken to be%
\begin{equation}
R_{jk}=\sum_{m=1}^{M}I(j\in m)\,\,I(int(B_{m})=k)\,\sqrt{var\,_{\mathbf{x}%
}[B_{m}(\mathbf{x})]}\text{.} \label{e31}%
\end{equation}
Here $I(\cdot)$ is an indicator of the truth of its argument, $m$ labels a
node of the $M$ node function tree, $B_{m}(\mathbf{x})$ is its corresponding
basis function, $int(B_{m})$ its interaction order and the variance is over
the distribution $p(\mathbf{x})$. This can be used as a filter to exclude
variables $x_{j}$ with small values of $R_{jk}$ (\ref{e31}) from the search
for $k$--variable interaction effects. Since node basis functions
$B_{m}(\mathbf{x})$ are not pure interaction effect functions (\ref{e22})
(\ref{e23}) the search for $n$--variable interaction effects should
include\ the union of all relevant variables with substantial contributions
(\ref{e31}) at that interaction level or higher.

In many applications the number of variables involved in higher order
interactions is less than that involved in main or lower order interaction
effects. Since computation increases very rapidly with the number of variables
examined, further variable reduction at the interaction level (\ref{e31}) can
substantially reduce computation even for minor variable reductions.

For the simulation example in Section \ref{sim} (below) exhaustive search over
all $30$ variables at each of four interaction levels requires computing
$2027970$ partial dependence functions with $2.03\times10^{12}$ corresponding
target function evaluations using (\ref{e15}) with $N=N_{\mathbf{z}}=1000$.
This would take approximately $36$ days on a Dell XPS 13 laptop. Approximating
the search by only including the ten variables estimated to be the most
important reduces the necessary number of partial dependence functions to
$15540$ requiring $1.55$ $\times$ $10^{10}$ target function evaluations. This
would take approximately seven hours.

Partial dependence screening using (\ref{e34}) identifies six interacting
variables reducing the number of partial dependence functions computed to
$1114$ with $1.11\times10^{9}$ target function evaluations using (\ref{e15}).
This reduces the computing time to approximately $30$ minutes. Building the
function tree for this example required $14.5$ seconds. Corresponding
interaction level screening using (\ref{e31}) eliminated all four--variable
interactions. Six variables remained for both the two--variable and
three--variable interaction searches and ten variables for main effects. This
requires a total of $384$ partial dependence functions. Employing (\ref{e17})
with $N=N_{z}=1000$ involved $4.14\times10^{5}$ target function evaluations
requiring $0.5$ seconds.

\subsection{Interaction investigation}

\begin{figure}[ptb]%
\centering
\includegraphics[
height=3.9868in,
width=4.2973in
]%
{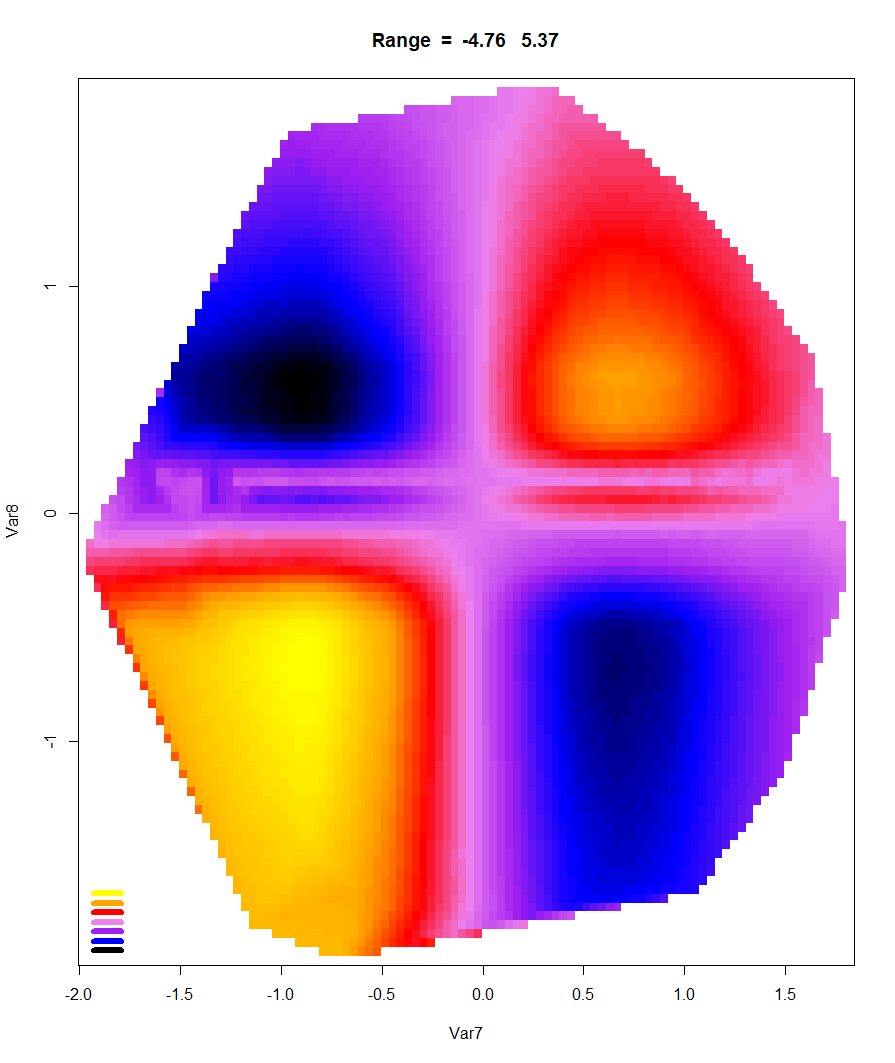}%
\caption{Interaction effect of variables $x_{7}$ and $x_{8}$ on simulated data
example of Section 3.}%
\label{fig5}%
\end{figure}

Once discovered, interactions as well as main effects can be visualized using
partial dependence plots. Figure \ref{fig5} shows a heat map representation of
the interaction effect between variables $x_{7}$ and $x_{8}$, $I(x_{7},x_{8})$
(\ref{e22}) (\ref{e23}), for the function tree model $\hat{F}(\mathbf{x})$
(Figs. 1 \& 2). Colors black, blue, purple, violet, red, orange, and yellow,
seen in the lower left legend, represent a continuum of increasing function
values from lowest ($-5.06$) to highest ($5.49$).%

One sees that joint high or joint low values of these two variables produce
relatively large increases in function value (red, yellow), whereas opposite
values, (low, high) or (high, low), give rise to large decreases (blue,
black). Since Fig. \ref{fig4} shows no higher order interactions involving
these variables, this function of $x_{7}$ and $x_{8}$ represents the
corresponding interaction effect associated with the target function estimate
$\hat{F}(\mathbf{x)}$, and does not reflect the nature of the data
distribution $p(\mathbf{x})$. One can verify this behavior by examining the
true target function (\ref{e77}).

Variables $x_{4}$ and $x_{5}$ are seen in Fig. \ref{fig4} to have a relatively
weak interaction. However, they are also seen to participate in a substantial
three--variable interaction with variable $x_{6}$ so that their two--variable
interaction function $I(x_{4},x_{5})$ \ (\ref{e22}) (\ref{e23}) may not
provide a complete description of their joint effect on the target estimate
$\hat{F}(\mathbf{x})$.%

\begin{figure}[ptb]%
\centering
\includegraphics[
height=6.269in,
width=6.0961in
]%
{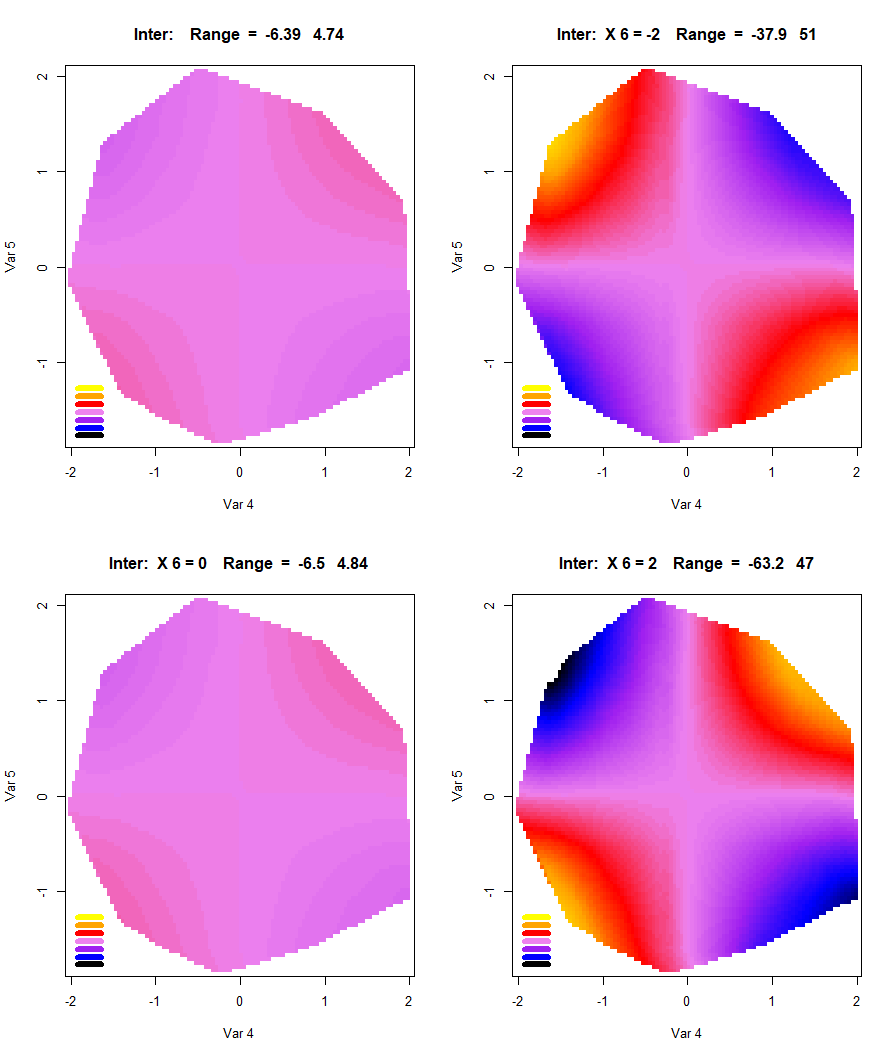}%
\caption{Average interaction effect $I(x_{4},x_{5})$ on variables $x_{4}$ and
$x_{5}$ (upper left), conditioned $I(x_{4},x_{5}|$ $x_{6})$ on $x_{6}=-2$
(upper right), $x_{6}=0$ (lower left) and $x_{6}=2$ (lower right) for
simulated data of Section 3.}%
\label{fig6}%
\end{figure}
Figure \ref{fig6} shows a plot of the average (unconditional) pure interaction
function $I(x_{4},x_{5})$ (upper left), and corresponding interaction
functions $I(x_{4},x_{5}\,|\,x_{6})$ conditioned at three values of $x_{6}%
\in\{-2,0,2\}$, all plotted on the same vertical scale ($-63,63$). Here a
black color represents the most negative values, yellow the most positive and
violet the smallest absolute values. The upper left frame shows that the
variation of unconditional interaction function $I(x_{4},x_{5})$ is quite
small ($\simeq$ violet) as indicated in Fig. \ref{fig4}. The lower left frame
shows the same for $I(x_{4},x_{5}\,|\,x_{6}=0)$. The two right frames however
show very strong interaction effects for $I(x_{4},x_{5}\,|\,x_{6}=-2)$ and
$I(x_{4},x_{5}\,|\,x_{6}=2)$. There is almost no interaction between $x_{4}$
and $x_{5}$ when $x_{6}=0$. For $x_{6}=\pm\,2$, there are strong but
\emph{opposite} interaction effects. This leads to an overall weak
two--variable interaction between $x_{4}$ and $x_{5}$ when averaged over the
distribution of $x_{6}$ (upper left). Seeing only this weak two--variable
interaction effect between ($x_{4},x_{5}$) without knowledge of the
corresponding three--variable $(x_{4,}x_{5},x_{6})$ interaction would lead to
the impression that these two variables influence the target $F(\mathbf{x})$
in an additive unrelated manner. As seen in the right two frames of Fig.
\ref{fig6} this is clearly not the case. Thus, in the presence of higher order
interactions, lower order representations can be misleading.

\section{Illustrations}

The function tree methodology described above was applied to a number of
popular public data sets to investigate the nature of the relationship between
the joint values of their predictor variables and outcome. In many cases the
corresponding function tree uncovered simple relationships involving additive
effects or at most a few two--variable interactions. Others were seen to
involve more complex structure. Examples of both are illustrated here.

The goal of the function tree approach is interpretational. As such its form
(\ref{e4}) (\ref{e5}) is somewhat more restrictive than other more flexible
methods focused purely on prediction accuracy such as XGBoost (Chen and
Guestrin 2016) or Random Forests (Breiman 2001). However, if on any particular
problem its accuracy is substantially less than these other methods the
quality of the corresponding interpretation may be questionable. For each of
the examples presented in this section root-mean-squared prediction error%
\begin{equation}
RMSE=\sqrt{\left.  \sum_{i=1}^{N}(y_{i}-\hat{F}(\mathbf{x}_{i}))^{2}\right/
\sum_{i=1}^{N}(y_{i}-\bar{y})^{2}} \label{e28}%
\end{equation}
is reported for default versions of the three methods: no tuning for Random
Forests and only model size tuning for function trees and XGBoost. In all
examples presented here the estimated interaction effects are based on partial
dependence functions (Section \ref{PDF}). Corresponding summaries using
partial association functions (Section \ref{PAF}) produced nearly identical
results in all cases.

\subsection{Capital bikeshare data\label{bike}}

This data set taken from the UCI Machine Learning Repository consists of 17379
hourly observations of counts of bicycle rentals in the Capital bikeshare
system in Washington, D. C. between 2011 and 2012. The outcome variable $y$ is
the rental count. The sixteen predictor variables are both categorical and
numeric consisting of corresponding time, weather and seasonal information.%

\begin{figure}[ptb]%
\centering
\includegraphics[
height=6.6668in,
width=5.7095in
]%
{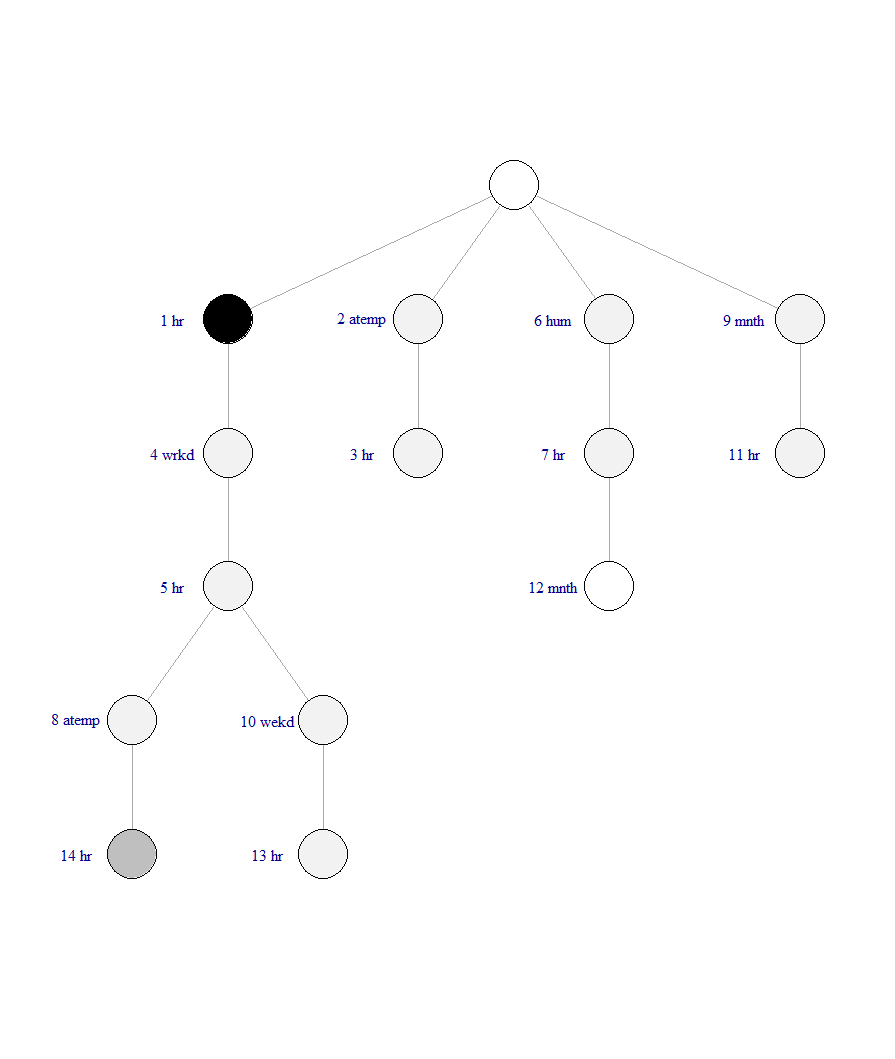}%
\caption{D. C. bicycle rental model function tree. }%
\label{fig7}%
\end{figure}
\begin{figure}[ptb]%
\centering
\includegraphics[
height=4.2291in,
width=5.8065in
]%
{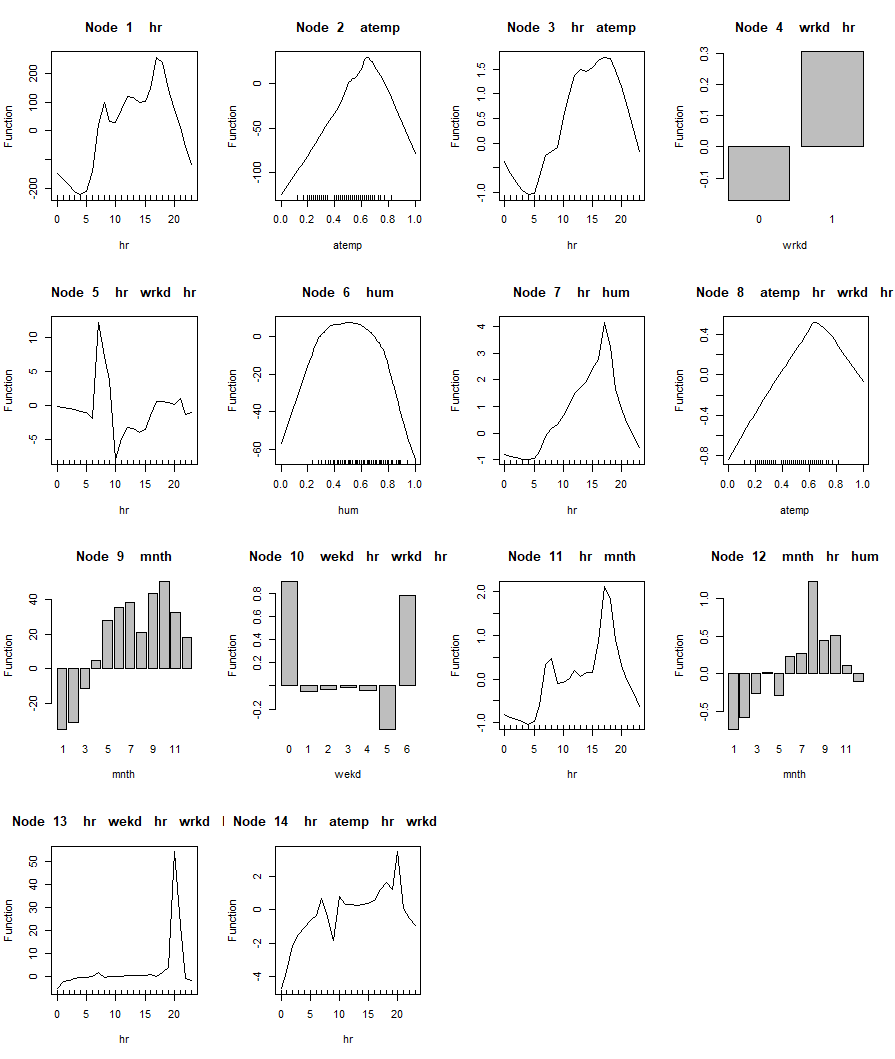}%
\caption{Node functions for bicycle rental function tree.}%
\label{fig8}%
\end{figure}
\begin{figure}[ptb]%
\centering
\includegraphics[
height=3.8564in,
width=5.4645in
]%
{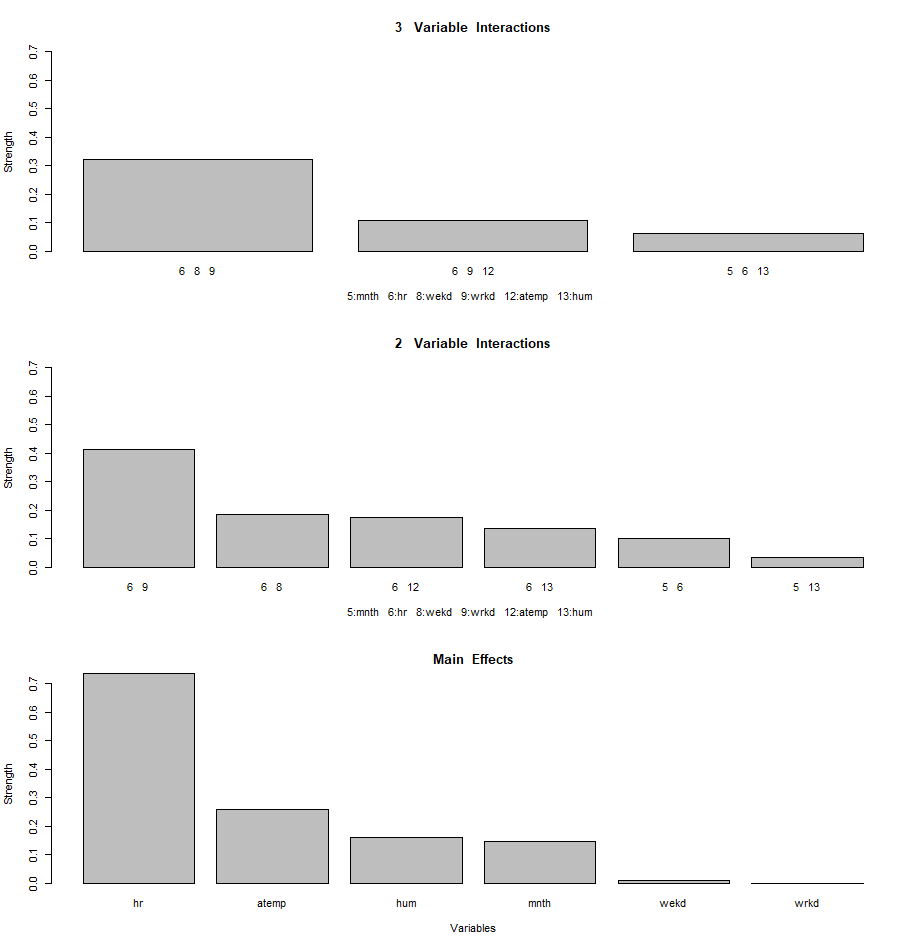}%
\caption{Interaction effect summary for bicycle rental function tree. }%
\label{fig9}%
\end{figure}

\begin{figure}[ptb]%
\centering
\includegraphics[
height=2.445in,
width=4.6135in
]%
{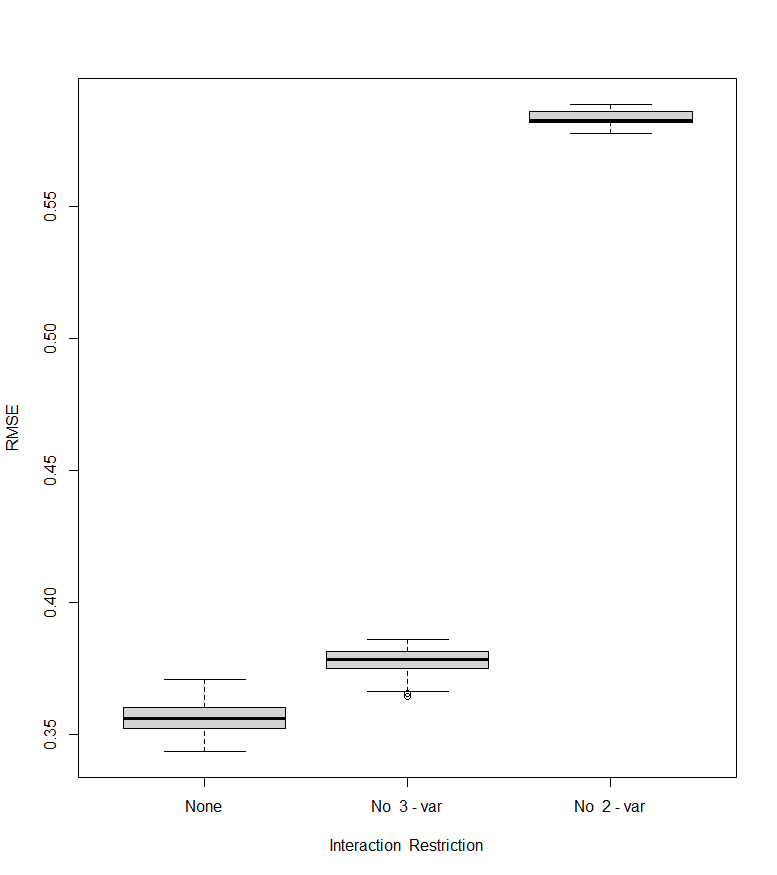}%
\caption{Boxplots of the distributions over 50 bootstrap replications of test
set root mean squared error of three function tree models on the bike rental
data: unconstrained (left), only main and two-variable interactions (middle),
and only main effects (right).}%
\label{figeaa}%
\end{figure}

Figure \ref{fig7} displays the tree for the function tree model and Fig.
\ref{fig8} shows the corresponding node functions. Its interaction effect
summary is shown in Fig. \thinspace\ref{fig9}. Figure \ref{fig9} indicates
existence of substantial two--variable and three--variable interactions. An
important question is the extent to which these effects are properties of the
target function and not just properties of a particular training sample. A way
to address this is through a bootstrap analysis (Efron 1979). Figure
\ref{figeaa} shows box plots of the distributions over fifty bootstrap
replications of test set root-mean squared-error (\ref{e28}) of function tree
models built under three constraints on maximum interaction order:
unconstrained (left), no three-variable interactions allowed (center), and no
interactions allowed (right). As seen, existence of both two--variable and
three--variable interactions are significant.%

\begin{figure}[ptb]%
\centering
\includegraphics[
height=3.1557in,
width=4.8526in
]%
{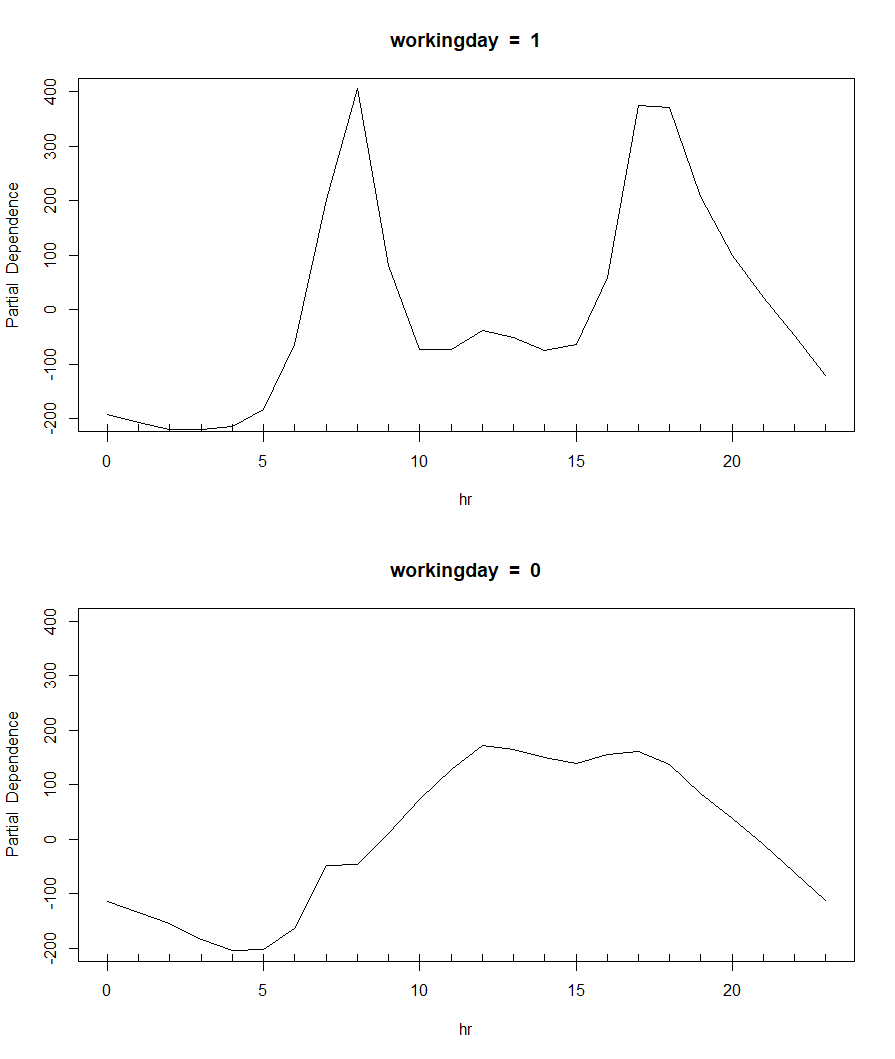}%
\caption{Partial dependence of DC bike rentals on time-of-day for working days
(top) and non working days (bottom).}%
\label{fig3f}%
\end{figure}

The largest estimated two--variable interaction effect is between variables
hour-of-day (\emph{hr}) and the binary indicator for working/non-working days
(\emph{wrk}). Figure \ref{fig3f} shows the partial dependence of DC bike
rentals on time-of-day conditioned \ on working days \emph{wrk }$=1$ (top) and
non working days \emph{wrk }$=0$ (bottom). They are seen to be quite different
indicating a strong interaction effect between these two variables. In
particular, the sharp reduction in rentals beginning at 8am and continuing
until 5pm on working days is seen not to exist on non-working days.%

\begin{figure}[ptb]%
\centering
\includegraphics[
height=4.8784in,
width=5.1059in
]%
{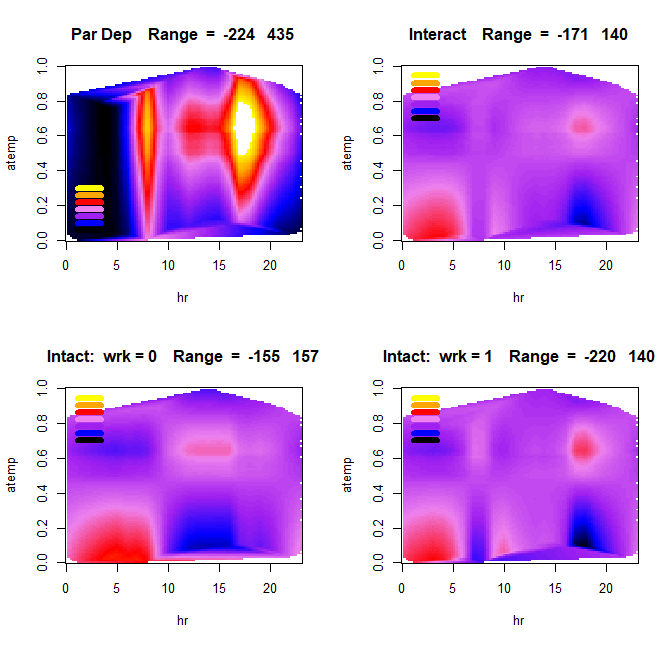}%
\caption{Partial dependence $PD(atemp,hr)$ of bike rentals jointly on adjusted
temperature and time-of-day (upper left), its pure interaction component
$I(atemp,hr)$ (upper right) and the pure interaction effect separately
conditioned on the value of working day $wrk=0$ (lower left) and $wrk=1$
(lower right)}%
\label{fig10}%
\end{figure}

The upper left frame of Fig. \ref{fig10} shows the partial dependence of bike
rentals jointly on adjusted temperature (\emph{atemp}) and time-of-day
(\emph{hr}), along with its pure interaction component $I($\emph{hr}%
$,$\emph{atemp}$)$ (top right). Figure \ref{fig9} indicates a three--variable
interaction between variables hour (\emph{hr}), working day indicator
(\emph{wrk}) \ and adjusted temperature (\emph{atemp}). Thus, the nature of
the interaction effect between \emph{hr} and \emph{atemp} may depend on the
value of \emph{wrk}. The bottom two frames of Fig. \ref{fig10} show this
interaction effect separately conditioned on the value of working day
$I($\emph{hr,atemp}$\,|\,$\emph{wrk}$=0)$ and $I($\emph{hr,atemp}%
$\,|\,$\emph{wrk}$=1)$. One sees substantial difference in the nature of this
interaction for the different values of \emph{wrk} reflecting the presence of
the three--variable interaction.%

\begin{figure}[ptb]%
\centering
\includegraphics[
height=3.8216in,
width=4.3223in
]%
{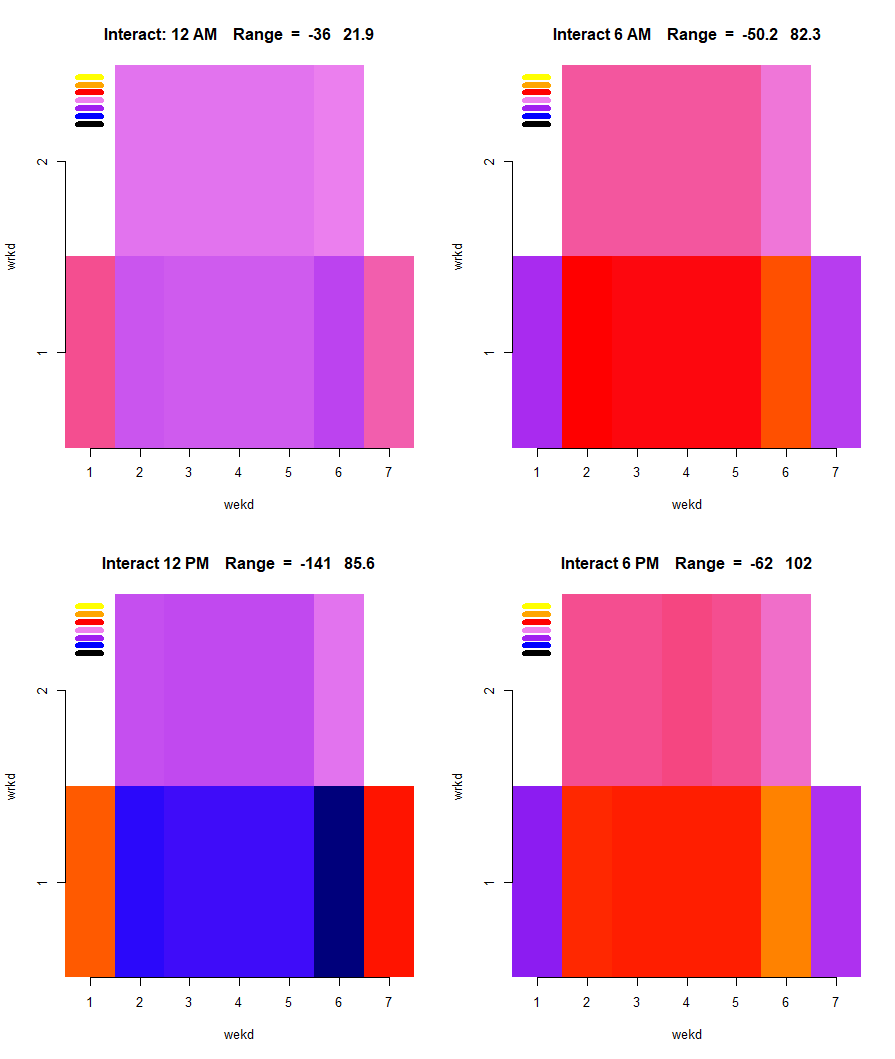}%
\caption{Bike data Interaction effect $I(wekd,wrk\,|\,hr)$ for $hr\in
\{midnight,6am,noon,6pm\}$.}%
\label{fig11}%
\end{figure}

Figure \ref{fig9} indicates that the strongest three--variable interaction
effect \thinspace involves the working day indicator (\emph{wrk}), day of the
week (\emph{wekd}) and hour of the day (\emph{hr}). This is illustrated in
Fig. \ref{fig11}. The interaction function $I($\emph{wekd,wrk\thinspace}%
%TCIMACRO{\TEXTsymbol{\vert}}%
%BeginExpansion
$\vert$%
%EndExpansion
\thinspace\emph{hr}) is displayed for \emph{hr\thinspace}$\in\,\{$%
\emph{midnight,6\thinspace am,noon,6\thinspace pm}$\}$. Note that there are no
observations for which Saturday (\emph{wekd }$=7$) and Sunday (\emph{wekd
}$=1$) are labeled as working days (\emph{wrk}$=2$). The absence of a
three--variable interaction would imply that $I($\emph{wekd,wrk\thinspace}%
%TCIMACRO{\TEXTsymbol{\vert}}%
%BeginExpansion
$\vert$%
%EndExpansion
\thinspace\emph{hr}$)$ is the same for all values of the variable \emph{hr}.
As seen in Fig. \ref{fig11} this is not the case especially for non working
weekdays at noon (lower left).

The $RSME$ (\ref{e28}) of function tree, XGBoost and Random Forest models on
this example was $0.34$, $0.37$ and $0.38$, respectively.

\subsection{Simulated data\label{sim}}

This example was presented in Hu \emph{et. al. }(2023). They generated data
using the target function%

\begin{align}
g(\mathbf{x)}  &  \mathbf{=}\sum_{j=1}^{5}x_{j}+0.5\sum_{j=6}^{8}\,x_{j}%
^{2}+\sum_{j=9}^{10}x_{j}\,\,I(x_{j}>0)+x_{1}\,x_{2}+x_{1}\,x_{3}+x_{2}%
\,x_{3}\label{e25}\\
&  +0.5\,x_{1}\,x_{2}\,\,x_{3}+x_{4}\,\,x_{5}+x_{4}\,x_{6}+x_{5}%
\,x_{6}+0.5\,I(x_{4}>0)\,\,x_{5}\,x_{6}\text{.}\nonumber
\end{align}
This is a function of ten variables involving numerous interactions involving
up to three variables. The response was simulated as%
\begin{equation}
y=g(\mathbf{x})+\varepsilon\text{, \ }\varepsilon\sim N(0,0.5^{2})\text{.}
\label{e27}%
\end{equation}
There were $30$ predictor variables. The first $20$ were simulated from a
multivariate Gaussian distribution with mean $0$, variance $1$ and equal
correlation $0.5$ between all pairs. Ten additional variables were included
that were independent of the first $20$ (irrelevant variables). They were also
simulated from a multivariate Gaussian distribution with mean $0$, variance
$1$ and equal correlation $0.5$ between all pairs. All predictors were
truncated to be within the interval $[-2.5,2.5]$. Here the sample size was
taken to be $N=20000$.%

\begin{figure}[ptb]%
\centering
\includegraphics[
height=3.7343in,
width=5.6481in
]%
{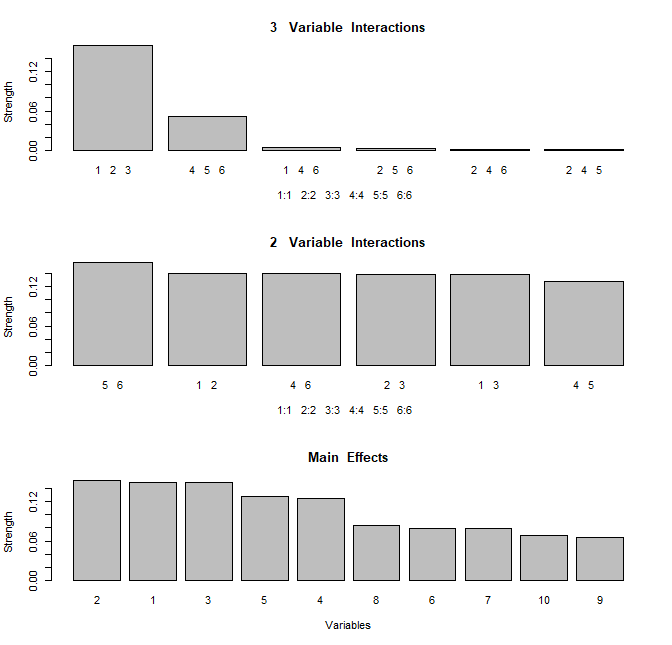}%
\caption{Strengths of main and interaction effects identified in the simulated
data example \ of Section 6.2.}%
\label{fig12}%
\end{figure}

Figure \ref{fig12} shows the largest estimated main and interaction effect
strengths up through four--variables. One sees that all of the main and
interaction effects represented in (\ref{e25}) are shown with substantial
strengths. Those not represented in (\ref{e25}), including all four--variable
interactions, either do not appear or do so with very low estimated strength.

The $RSME$ (\ref{e28}) of function tree, XGBoost and Random Forest models on
this example was $0.062$, $0.147$ and $0.174$ respectively.

\subsection{Pumadyn data}

The other examples were chosen to represent data with somewhat complex target
function structure. Here we illustrate on a data set where the function tree
uncovers very simple structure. The data (Corke 1996) is produced from a
realistic simulation of the dynamics of a Puma 560 robot arm. The task is to
predict the angular acceleration of one of the robot arm's links. There are
8192 observations involving 32 input variables that include angular positions,
velocities and torques of the robot arm.%

\begin{figure}[ptb]%
\centering
\includegraphics[
height=3.6222in,
width=5.6513in
]%
{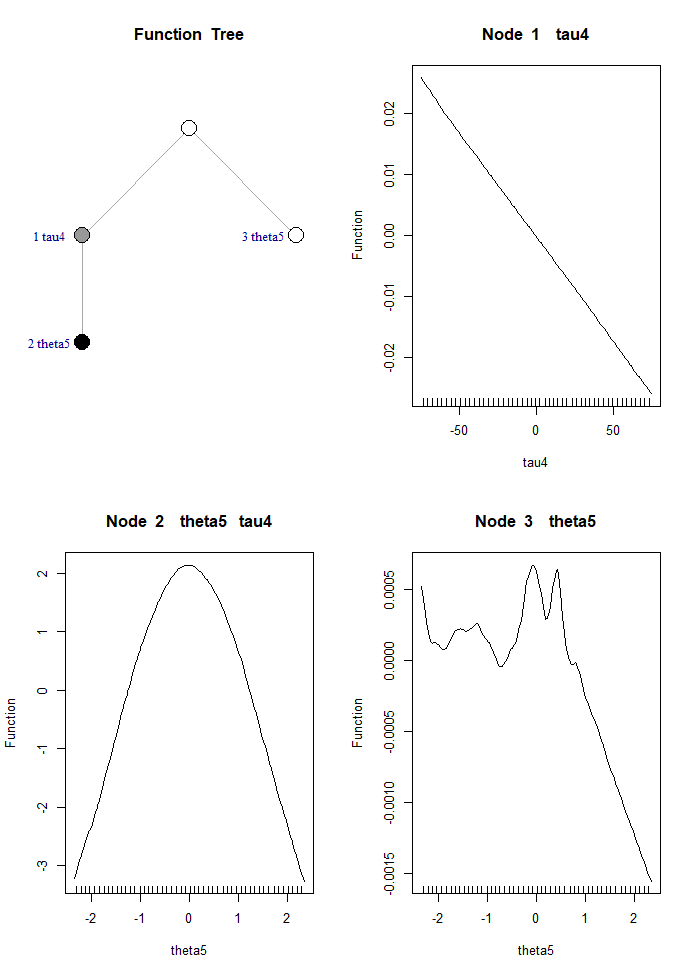}%
\caption{Function tree model resulting from the Pumadyn data with the tree
(upper left) and respective node functions.}%
\label{fig13}%
\end{figure}
\begin{figure}[ptb]%
\centering
\includegraphics[
height=3.4774in,
width=4.2229in
]%
{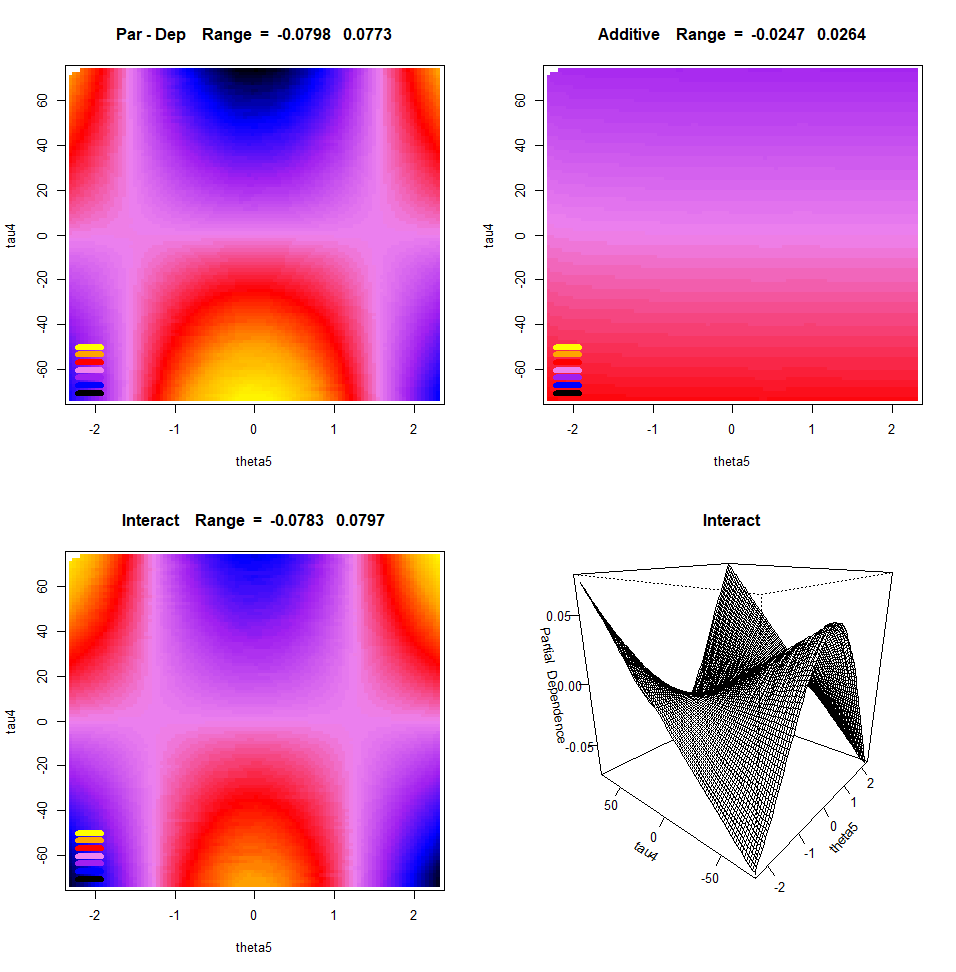}%
\caption{Pumadyn data: partial dependence on variables \emph{tau4} and
\emph{theta5 }(top left), additive contribution (top right), pure interaction
(bottom left) and pure interaction perspective mesh plot (bottom right).}%
\label{fig14}%
\end{figure}

Figure \ref{fig13} shows the resulting entire function tree model. The upper
left frame shows the tree and the other frames show the node functions. Figure
\ref{fig14} shows the (only) interaction effect between \emph{tau4} and
\emph{theta5}. The upper left frame shows the partial dependence function
$PD(theta5,tau4)$; upper right frame shows its additive component and lower
left frame its pure interaction effect $I(theta5,tau4)$. The lower right frame
represents $I(theta5,tau4)$ as a perspective mesh plot. Of the 32 input
variables, the function tree model selected only two of them to explain 97\%
of the variance of the target.

The $RSME$ (\ref{e28}) of function tree, XGBoost and Random Forest models on
this example was $0.18$, $0.21$ and $0.27$, respectively.

\subsection{SGEMM GPU kernel performance\label{gpu}}

This data set from the UCI Machine Learning Repository measures the running
time required for a matrix-matrix product using a parameterizable SGEMM GPU
kernel. There are 241600 observations with 14 predictor variables:

\begin{center}
SGEMM GPU kernel performance predictor variables

\qquad$%
\begin{array}
[c]{lll}%
\text{1-2} & \text{\emph{mwg,\thinspace nwg}} & \text{per-matrix 2Dtiling at
workgroup level}\\
\text{3} & \text{\emph{kwg}} & \text{inner dimension of 2Dtiling at workgroup
level}\\
\text{4-5} & \text{\emph{mdimc,\thinspace ndimc}} & \text{local workgroup
size}\\
\text{6-7} & \text{\emph{mdima,\thinspace ndimb}} & \text{local memory
shape}\\
\text{8} & \text{\emph{kwi}} & \text{kernel loop unrolling factor}\\
\text{9-10} & \text{\emph{vwm,\thinspace vwn}} & \text{per-matrix vector
widths for loading and storing}\\
\text{11-12} & \text{\emph{strm,\thinspace strn}} & \text{enable stride for
accessing off-chip memory within a single thread}\\
\text{13-14} & \text{\emph{sa,\thinspace sb}} & \text{per-matrix manual
caching of the 2D workgroup tile.}%
\end{array}
$
\end{center}

\begin{figure}[ptb]%
\centering
\includegraphics[
height=3.685in,
width=5.6567in
]%
{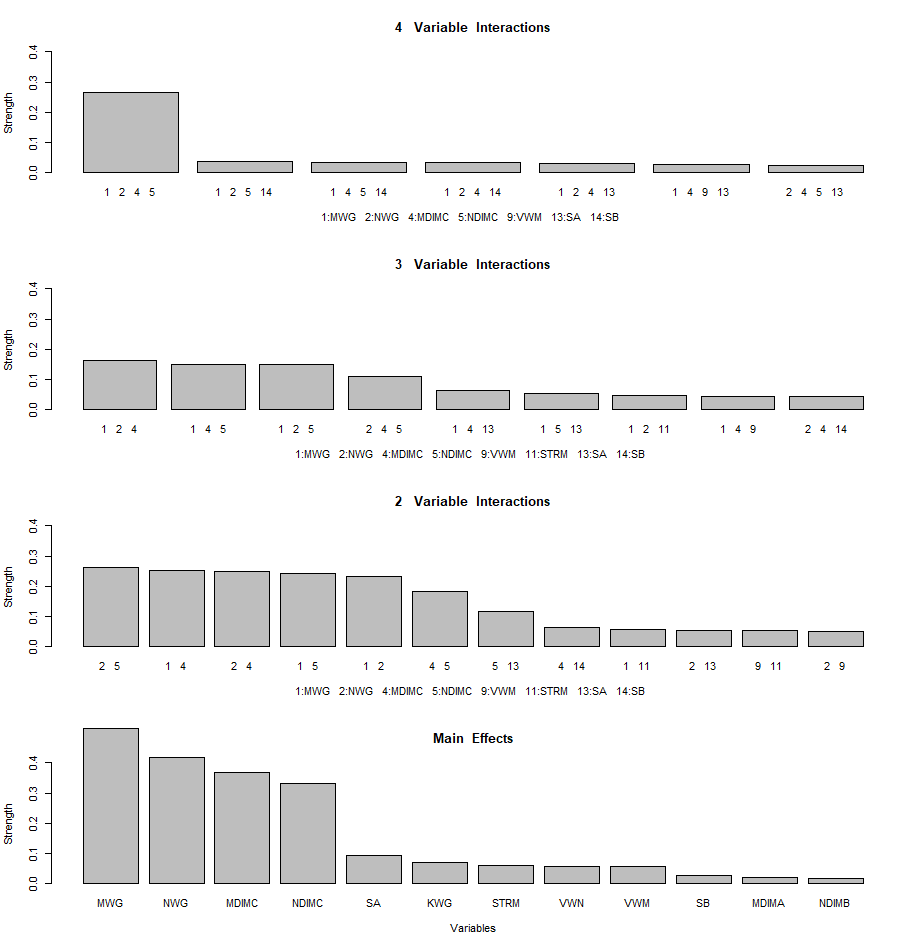}%
\caption{Strengths of main and interaction effects identified in the SGEMM GPU
kernel performance data.}%
\label{fig15}%
\end{figure}

As suggested by the authors, the outcome was taken to be the logarithm of the
running time. All predictor variables realize at most four distinct values and
were therefore treated as being categorical. Figure \ref{fig15} shows the
estimated interaction effect profile for this data. Among the larger effects
are four main, seven two--variable, four three--variable and a large
four--variable interaction effect.%

\begin{figure}[ptb]%
\centering
\includegraphics[
height=2.8781in,
width=5.7285in
]%
{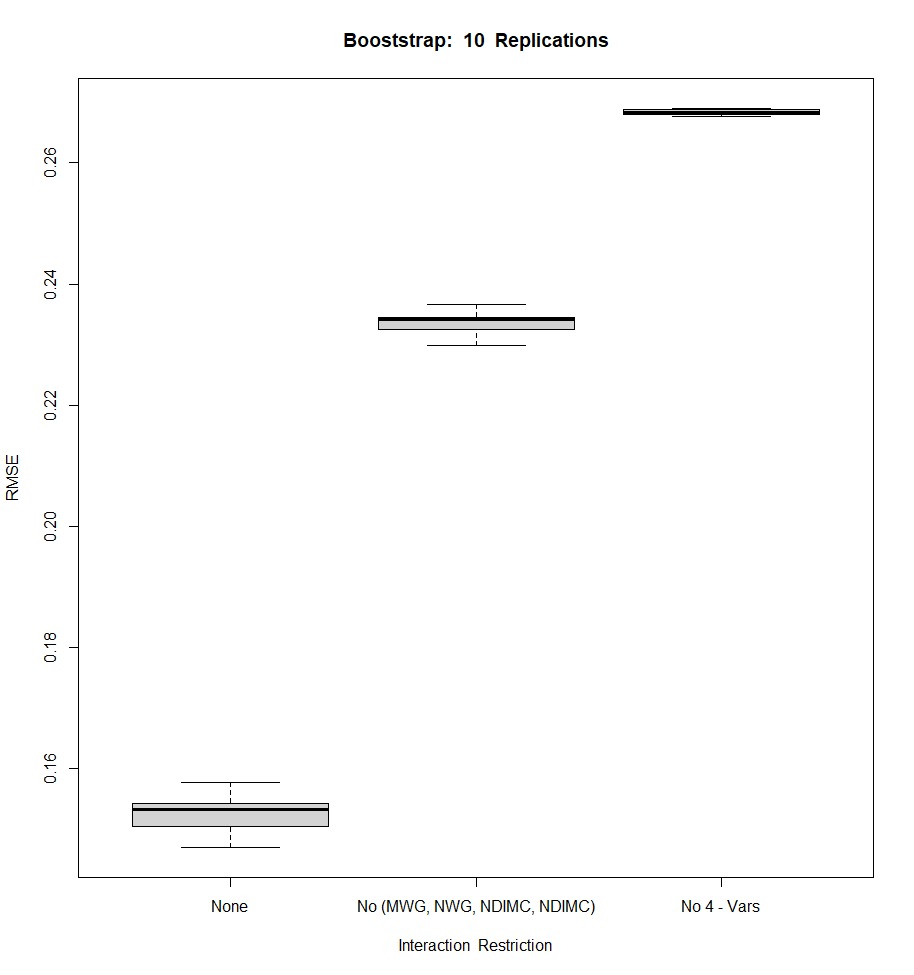}%
\caption{Boxplots of the distributions over 10 bootstrap replications of test
set root mean squared error of three function tree models on the GPU
performance data: unconstrained (left), no (\emph{mwg,nwg,mdimc,ndimc})
interaction (center), and no four variable interactions (right)}%
\label{fig16}%
\end{figure}
Figure \ref{fig16} shows box plots of the distributions over ten bootstrap
replications of test set root-mean squared-error (\ref{e28}) of function tree
models built under three interaction constraints: unconstrained (left),
four--variable interaction among variables (\emph{mwg,nwg,mdimc,ndimc})
prohibited (center), and all four--variable interactions prohibited (right).
Note that not allowing the four--variable interaction among
(\emph{mwg,nwg,mdimc,ndimc}) does not preclude any of these variables from
participating in lower order interactions, or four--variable interactions with
other variables. One sees that incorporating the (\emph{mwg,nwg,mdimc,ndimc})
four--variable interaction is very important to model accuracy as are other
four--variable interactions to a lesser extent.%

\begin{figure}[ptb]%
\centering
\includegraphics[
height=6.7205in,
width=6.1116in
]%
{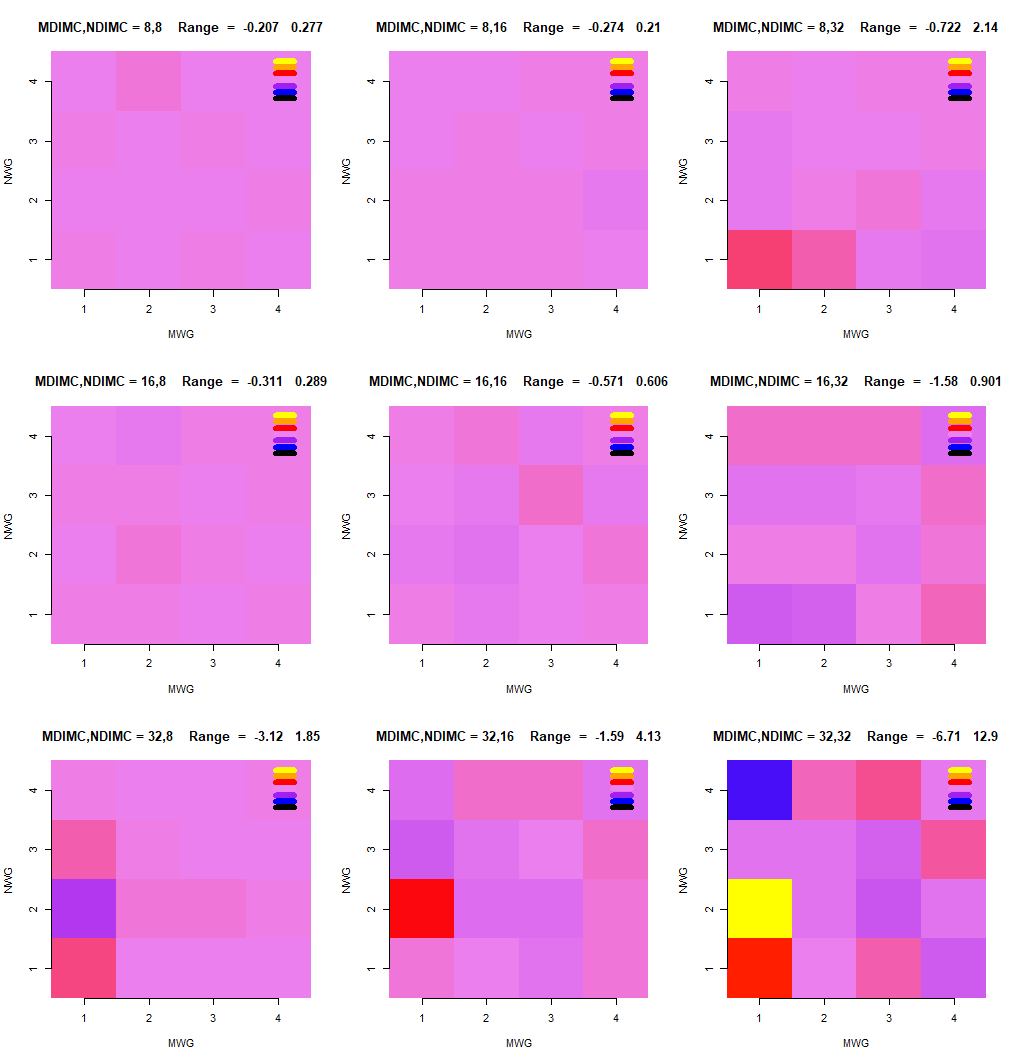}%
\caption{ SGEMM GPU kernel interaction effect differences with and without the
$($\emph{mwg}$,$\emph{nwg}$\,,$\emph{mdimc}$,$\emph{ndimc}$)$ four--variable
interaction effect. }%
\label{fig21}%
\end{figure}

Since the (\emph{mwg,nwg,mdimc,ndimc}) interaction effect is inherently four
dimensional it cannot be completely represented by a two dimensional display.
Unlike three--variable interactions it cannot be represented by a series of
two dimensional plots conditioned on the value of a third variable. However it
is possible to gain some insight about this four--variable interaction effect
through graphical methods.

Figure \ref{fig21} depicts the difference between interaction effects computed
from two models%
\begin{equation}
\hat{D}(\mathbf{x})=\hat{F}(\mathbf{x})-\hat{G}(\mathbf{x})\text{.}
\label{e32}%
\end{equation}
The first $\hat{F}(\mathbf{x})$ is the full model with all interactions
allowed (Fig. \ref{fig16} left). The second model $\hat{G}(\mathbf{x})$ was
built under the constraint that no four--variable interaction effect among
variables (\emph{mwg,nwg,mdimc,ndimc}) was allowed (Fig. \ref{fig16} center).
Thus, any structure detected in $\hat{D}(\mathbf{x})$ is due to presence of
this four--variable interaction.

Figure \ref{fig21} shows nine bivariate interaction effects differences
between these two models (\ref{e32})%
\[
\emph{I}_{\hat{D}}(\text{\emph{mwg,nwg\thinspace%
%TCIMACRO{\TEXTsymbol{\vert}}%
%BeginExpansion
$\vert$%
%EndExpansion
\thinspace mdimc}}\,\in\,\{\emph{8,16,32}\}\,\,\times
\,\,\text{\emph{ndimc\thinspace}}\in\,\{\emph{8,16,32}\})
\]
where \emph{mwg}$,$\emph{nwg }$\in\{\emph{16,32,64,128}\}$. For the top two
rows \emph{mdimc}$\,\in\,\{8,16\}$, the model differences are seen to be small
in absolute value ($\simeq$ violet) for all joint values of the other
variables, indicating that the four--variable interaction has little to no
effect for those collective values. For the last row (\emph{mdimc\thinspace
\thinspace}$=32$) one sees that the absolute differences are larger (red,
yellow, blue), especially for \emph{ndimc\thinspace\thinspace}$=32$\emph{.}
This four--variable interaction effect is seen to be mainly the result of a
very large increase in (log) running time for the GPU matrix product when
simultaneously \emph{mdimc }and \emph{ndimc }realize their largest values
while at the same time \emph{mwg }and\emph{ nwg }are at their smallest values.

The $RSME$ (\ref{e28}) of function tree, XGBoost and Random Forest models on
this example was $0.12$, $0.09$ and $0.16$, respectively.

\subsection{Global surrogate}

Function trees can be used as interpretable surrogate models to investigate
the predictions of black box models. One simply uses the black box predictions
along with the corresponding predictor variables as input. To the extent that
the function tree fit is accurate, all of its interpretational tools can then
be applied to study the nature of the black box model. This is illustrated
using the simulated data of Hu \emph{et. al. }(2023), shown in Section
\ref{sim}, where the true underlying structure (\ref{e25}) is known.%

\begin{figure}[ptb]%
\centering
\includegraphics[
height=3.5483in,
width=5.169in
]%
{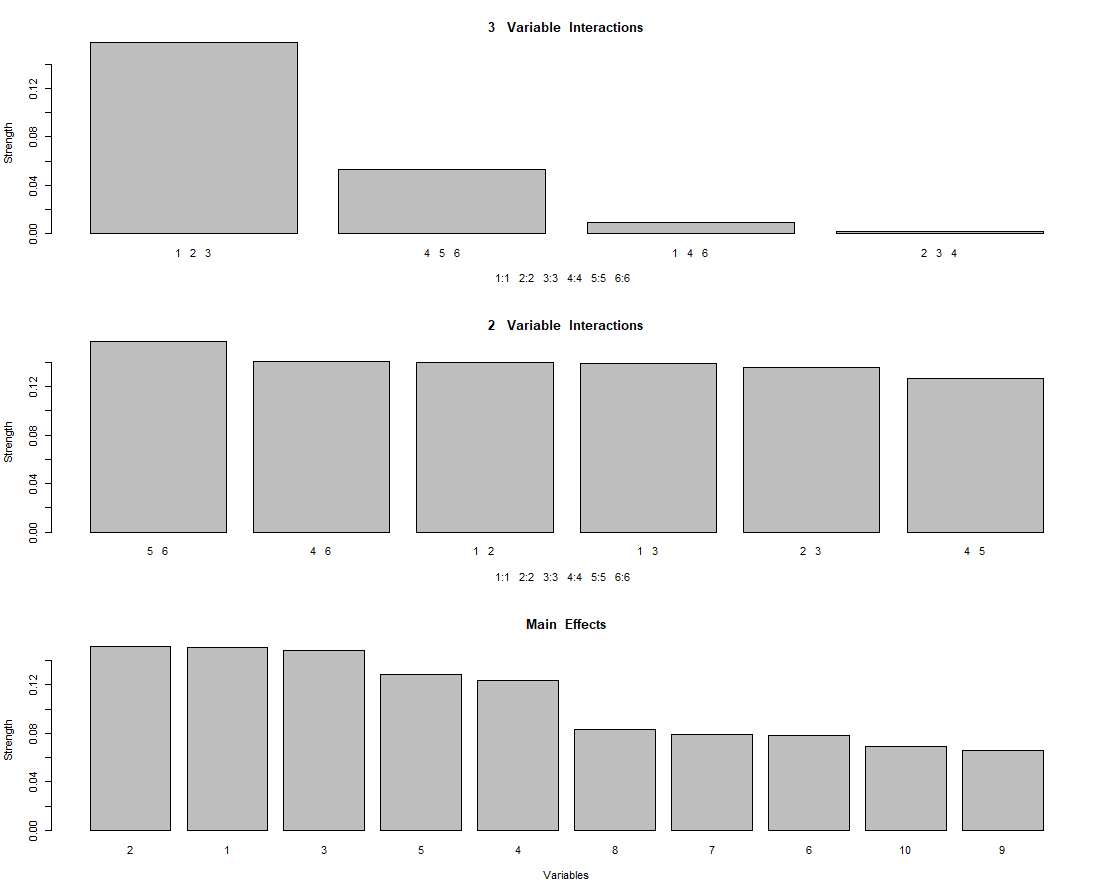}%
\caption{Strengths of main and interaction effects identified in the XGBoost
regression model on the simulation data of Section 6.2}%
\label{fig17}%
\end{figure}
\begin{figure}[ptb]%
\centering
\includegraphics[
height=3.2967in,
width=5.2702in
]%
{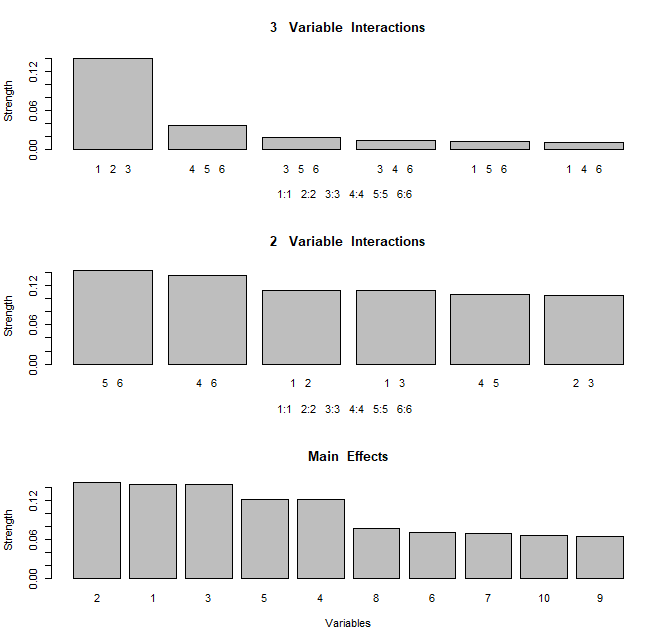}%
\caption{Strengths of main and interaction effects identified in the Random
Forest regression model on the simulation data of Section 6.2}%
\label{fig20}%
\end{figure}

First XGBoost is applied in regression mode to model the data (\ref{e27})
producing an estimate $\hat{g}(\mathbf{x})$. The resulting test data
root--mean--squared error%
\begin{equation}
RMSE(\hat{g})=\left\{  \hat{E}_{\mathbf{x}}[(g(\mathbf{x})-\hat{g}%
(\mathbf{x}))^{2}]/Var(g(\mathbf{x}))\right\}  ^{\frac{1}{2}} \label{e29}%
\end{equation}
was $RMSE(\hat{g})=0.14$. The output of the XGBoost model $\hat{g}%
(\mathbf{x})$ is then fit with a function tree. The $RMSE$ for this fit was
$0.13$ indicating that the function tree accurately reflects the XGBoost
model. Figure \ref{fig17} displays the resulting largest main and interaction
effects detected by the function tree in the XGBoost model. This can be
compared to the function tree based on the original data \ shown in Fig.
\ref{fig12}, as well as to the structure of the true target function
(\ref{e25}). One sees that the XGBoost regression model here closely captures
the true target function structure as reflected in its function tree representation.

Next a Random Forest is applied to the same data. The resulting test data
root--mean--squared error (\ref{e29}) was $RMSE(\hat{g})=0.17$. The output of
this random forest model was fit by a function tree producing a corresponding
$RMSE$ of $0.053$. Figure \ref{fig20} shows the largest main and interaction
effects detected in the Random Forest model. While capturing the nature of the
target function reasonably well, the random forest appears to somewhat under
estimate the $(x_{4},x_{5},x_{6})$ interaction strength and identify several
spurious three--variable interactions.%

\begin{figure}[ptb]%
\centering
\includegraphics[
height=3.6313in,
width=5.188in
]%
{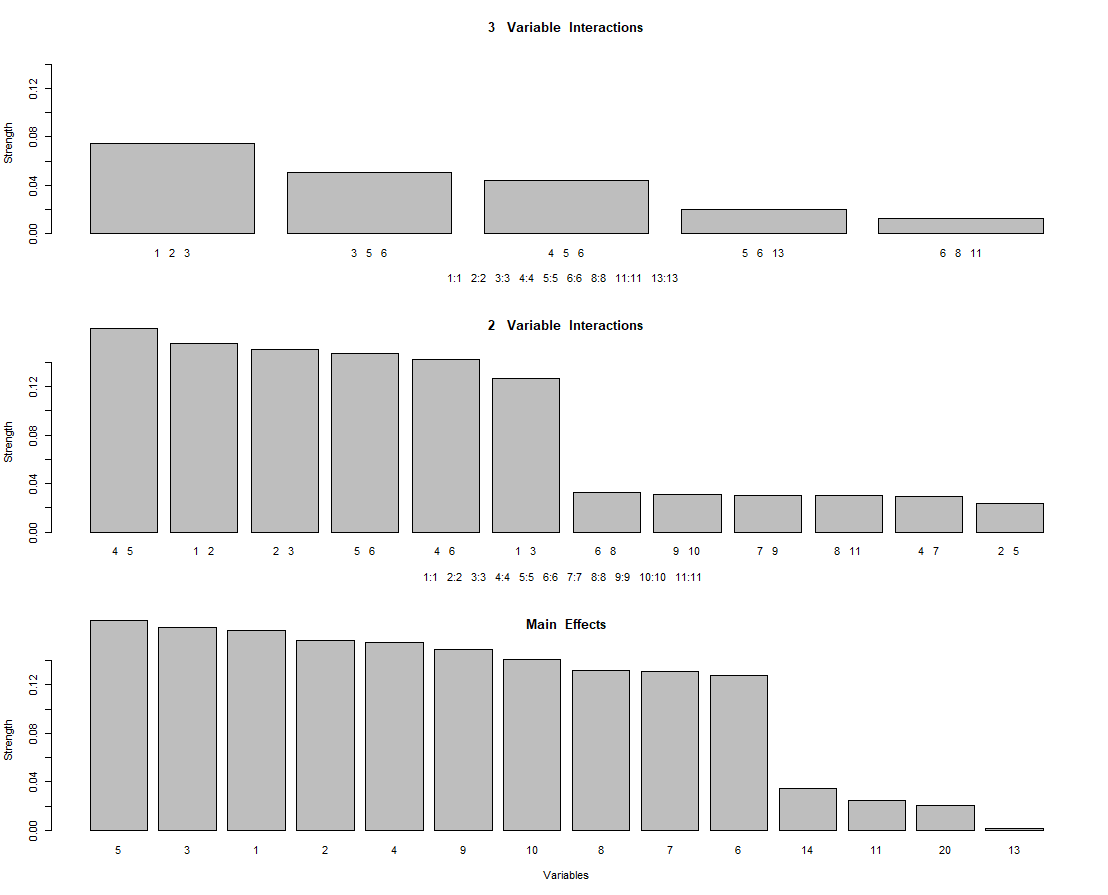}%
\caption{Strengths of main and interaction effects identified in the XGBoost
classification model on the simulation data of Section 6.2}%
\label{fig18}%
\end{figure}

Finally here we apply the surrogate approach in a classification setting. The
outcome variable is binary, $y\in\{0,1\}$ with $\Pr(y=1\,|\,\mathbf{x}%
)=1/(1+\exp(-g(\mathbf{x}))$. The log--odds function $g(\mathbf{x})$ is given
by (\ref{e25}). Logistic XGBoost was applied to this $(\mathbf{x},y)$ data
producing a log--odds estimate $\hat{g}(\mathbf{x})$. The resulting
root-mean-squared-error $RMSE(\hat{g})=0.39$ (\ref{e29}) is in this case much
larger owing to the loss of information associated with providing the learning
algorithm with only the sign of the outcome rather than its actual numerical
value. This logistic XGBoost model was then fit with a function tree with
resulting $RMSE$ of $0.19$. Although larger than in the regression case this
still represents a fairly close fit ($R^{2}=0.96$). The result is shown in
Fig. \ref{fig18}. As would be expected this lower accuracy logistic XGBoost
model less perfectly captures the true target function (\ref{e25}). It
uncovers all of the true main and interaction effects. It also indicates a
number of spurious main and two--variable interactions at somewhat lower
strength. The strength of the interaction among variables $(x_{1},x_{2}%
,x_{3})$ is under estimated and there are several spurious three--variable
interactions indicated at comparable strength. In spite of its lower accuracy
the logistic XGBoost model is still seen in Fig. \ref{fig18} to reflect much
of intrinsic structure of the target log--odds $g(\mathbf{x})$ (\ref{e25}).

\section{Previous work}

The origins of the function tree approach lie with the additive modeling
procedure of Hastie and Tibshirani (1990). They used nonparametric univariate
smoothers and backfitting to produce models with no interactions. Function
trees can be viewed as generalizing that method to discover and include
unspecified interaction effects.

The closest predecessor to function trees is MARS (Friedman 1991). It models a
general multivariate function as a sum of products of simple basic univariate
functions of the predictor variables and can in principle model interaction
effects to high order. The basic univariate function used with MARS is a dReLU
(double ReLU)%
\begin{equation}
d(x)=a\cdot(t-x)_{+}+b\cdot(x-t)_{+} \label{e30}%
\end{equation}
characterized by two slope parameters $(a,b)$ and knot location $t$. The MARS
forward stepwise modeling strategy is similar to that used with the function
tree approach.

The principal difference between MARS and function trees involves their basic
building blocks. MARS employs the simple dReLU (\ref{e30}) at every step.
Function trees employ more general functions of the predictor variables
$f(x_{j})$\ estimated by user selected smoothers. This allows customizing of
the estimation method to the nature of each individual predictor variable
distribution. It also produces more parsimonious models (smaller trees) since
many dReLUs (\ref{e30}) can be required to approximate even simple whole curves.

A second difference with MARS is that function trees allow different functions
of the \emph{same} variable to appear multiple times in the products of a
single basis function (\ref{e5}), as seen in Fig. \ref{fig7}. This provides
more flexibility by incorporating stepwise multiplicative as well as additive
modeling. Besides increased accuracy this further reduces model size. In
addition to being more interpretable, smaller models allow faster computation
of their partial dependence functions.

The use of partial dependence functions to uncover interaction effects was
proposed by Friedman and Popescu (2008). The innovation here is their rapid
identification and computation afforded by the function tree representation.
This allows a comprehensive search for all interaction effects up to high
orders. Other techniques that have been proposed for interaction detection are
mostly based on the functional ANOVA decomposition (see Roosen 1995, Hooker
2004 \& 2007, Kim \emph{et. al. }2009, Lengerich \emph{et. al. }2020, Hu
\emph{et. al. }2023, and Walters \emph{et. al. }2023). So far, computational
considerations have limited these approaches to two--variable interactions.
Lou \emph{et. al.} (2013) use a bivariate partitioning method to screen for
two--variable interactions. Tang, \emph{et. al. }(2016) combine the functional
ANOVA technique with the polynomial dimensional decomposition to reduce
computation with independent variables.

\section{Discussion}

While sometimes competitive in accuracy with other more flexible methods such
as XGBoost and Random Forests, the focus of the function tree approach is on
interpretation. The goal is to provide a representation of the target function
that exposes its interaction effects and provides a framework for their rapid
calculation, especially those involving more than two variables. Almost all
research into interaction detection to date has been limited to that involving
just two variables. In fact, in many settings the unqualified term
\textquotedblleft interaction effect\textquotedblright\ is meant to refer to
two variables only.

Focusing only on two--variable interactions is natural because it reduces the
size of the search and once interactions are detected their functional forms
are easily examined by traditional graphical methods such as heat maps,
contour plots or perspective mesh plots, etc. Higher order interactions
involve more variables and their higher dimensional structures are not as
easily represented. As illustrated in Figs. \ref{fig6}, \ref{fig10} and
\ref{fig11}, three--variable interactions can be visualized by viewing a
series of bivariate interaction plots conditioned on selected values of
another variable. As seen in Fig. \ref{fig21} interactions involving more
variables can be investigated by contrasting models with and without those
interactions included.

In addition to their direct interpretational value, knowledge of the existence
and strengths of higher order interactions can be important as they place
interpretational limits on the nature of those of lower order involving the
same variables. As noted above, the functional form of an interaction effect
involving variables $\{x_{j}\}_{1}^{n}$ depends on the value of another
variable, say $x_{k}$, if there exists a higher order interaction involving
both $\{x_{j}\}_{1}^{n}$ and $x_{k}$. If there are no such substantial higher
order interactions, the functional form of the $\{x_{j}\}_{1}^{n}$ interaction
is well defined and represents the isolated joint contribution of those
variables to the the target function. If such higher order interactions do
exist then the form of the $\{x_{j}\}_{1}^{n}$ interaction is not well defined
as it depends on the value of $x_{k}$ and its corresponding functional form
becomes an average over the joint distribution of all such variables. As seen
for example in Fig. \ref{fig6}, this can lead to highly misleading
interpretations. Tan\emph{ et. al.} (2023) discuss problems interpreting lower
order effects in the presence of higher order interactions.

In applications involving training data with very large absolute correlations
among subsets of predictor variables, main and interaction effects of various
levels involving those variables tend not to be separable. This can cause
substantial spurious interaction effects to be reported. These can be detected
by comparing the lack-of-fit (\ref{e28}) for models constructed with and
without the questionable effects included, as illustrated in Figs.
\ref{figeaa} and \ref{fig16}.

A popular interpretational tool used to investigate predictive models is a
measure of the impact or importance of the respective individual input
variables on model predictions. There are a wide variety of definitions for
variable importance each providing different information. See Molnar (2023)
for a comprehensive survey. In the presence of interactions the contribution
of a given predictor variable depends on the values of the other variables
with which it interacts. Interaction effect summaries (e.g. Figs. \ref{fig9},
\ref{fig12} and \ref{fig15}) are thus more comprehensive than corresponding
variable importance summaries. Variable importances can be derived from
interaction based functional decompositions. For example, Gevaert and Saeys
(2022) and Walters \emph{et. al.} (2023) use them to derive Shapley (1953) values.

In the examples presented here partial dependence functions were used to both
detect and examine interaction effects. Computational considerations largely
dictate their use for the former. However, once uncovered, identified main and
interaction effects can be examined by any appropriate method. For example,
accumulated local effects (ALE) functions (Apley and Zhu 2020) can be employed
for this purpose. Also, partial dependence based search results can be used to
guide methods for constructing functional ANOVA decompositions.

\end{document}